\def\arxivpreprint{1}
\documentclass{article}
\usepackage{iclr2027_conference,times}

\usepackage{amsmath,amsfonts,bm}

\def\eqref#1{equation~\ref{#1}}
\def\1{\bm{1}}

\DeclareMathAlphabet{\mathsfit}{\encodingdefault}{\sfdefault}{m}{sl}
\SetMathAlphabet{\mathsfit}{bold}{\encodingdefault}{\sfdefault}{bx}{n}

\usepackage{amssymb}
\usepackage{graphicx,booktabs,tabularx,float}
\usepackage{wrapfig}
\usepackage{placeins}
\usepackage{needspace}
\usepackage{fvextra}

\newlength{\figurecaptionspace}
\AddToHook{env/figure/begin}{\setlength{\abovecaptionskip}{\figurecaptionspace}\setlength{\belowcaptionskip}{0pt}}
\AddToHook{env/figure*/begin}{\setlength{\abovecaptionskip}{\figurecaptionspace}\setlength{\belowcaptionskip}{0pt}}

\newlength{\tablecaptiontopspace}
\AddToHook{env/table/begin}{\setlength{\abovecaptionskip}{\tablecaptiontopspace}\setlength{\belowcaptionskip}{6pt}}
\AddToHook{env/table*/begin}{\setlength{\abovecaptionskip}{\tablecaptiontopspace}\setlength{\belowcaptionskip}{6pt}}

\usepackage{hyperref}
\usepackage{url}

\title{Probing Local Conditional Fit for Diffusion Model Membership Inference}

\author{%
ByungHee Cha \\
Sony Group Corporation \\
KAIST \\
\And
Naoki Murata \\
Sony Group Corporation \\
\And
Yuhta Takida \\
Sony Group Corporation \\
\And
Jong Chul Ye\thanks{Corresponding authors.} \\
KAIST \\
\And
Yuki Mitsufuji\footnotemark[2] \\
Sony Group Corporation
}

\ifdefined\arxivpreprint
  \iclrfinalcopy
\fi
\begin{document}
\raggedbottom

\setcounter{footnote}{1}
\maketitle
\ifdefined\arxivpreprint
  \lhead{Preprint}
\fi

\begin{abstract}
Membership inference determines whether a candidate example was used to train a text-to-image diffusion model and is an important tool for auditing training-data use. We identify a directional membership signal in \emph{conditional prediction shifts}, the changes in predicted noise under alternative text conditions at a fixed noisy input: whether movement toward the alternative prediction initially lowers or raises the denoising loss from its value under the original caption. Members exhibit less room for such local improvement. To quantify this response, we propose \emph{LoFIT} (Local Conditional Fit), which measures the initial response of the denoising loss along normalized prediction shifts and aggregates it across multiple text conditions and diffusion states. LoFIT requires no target-model gradients and can be computed using only forward queries to the denoiser. Evaluations on four datasets show improved membership discrimination across fine-tuned and pretrained models. LoFIT distinguishes members early in fine-tuning, while the naive denoising-loss baseline remains near chance. Controlled comparisons show that the membership signal is concentrated in the first-order response, with little discrimination from prediction-shift magnitude alone.
\end{abstract}

\section{Introduction}
\label{sec:introduction}

Text-to-image diffusion models are trained on large collections of paired images and captions~\citep{rombach2022latent} and can memorize and reproduce training images~\citep{carlini2023extracting,somepalli2023replication,hong2024membenchmemorizedimagetrigger}. Because these collections may contain private, copyrighted, or otherwise unauthorized material, determining whether a candidate example was used in training is important for auditing data provenance and training-data use. Membership inference attacks (MIAs) determine whether a queried example participated in training~\citep{shokri2017membership} and can provide such audits for text-to-image diffusion models~\citep{zhai2024clid,qi2026sdmia}. We study a gray-box setting in which an auditor can query the noise predictor at chosen noisy inputs, timesteps, and text conditions, but cannot inspect model weights or obtain target-model gradients. This access enables direct evaluation of conditional denoising fit without differentiating through the target model.

Several diffusion-model MIAs use denoising or reconstruction errors as their primary membership signals~\citep{duan2023secmi,kong2024pia,tang2024quantile}. Denoising loss is a natural starting point not only because it is the training objective but also because it is connected to likelihood through the variational bound~\citep{ho2020ddpm}. Generalization allows diffusion models to denoise unseen images accurately, making absolute denoising error alone an incomplete indicator of membership~\citep{zhai2024clid}. CLiD~\citep{zhai2024clid} extends this view to the relationship between an image and its text condition. Motivated by conditional overfitting, it compares losses under the original caption and modified conditions. Its results suggest that membership information can reside in the learned image-text relationship, beyond denoising accuracy under a single condition.

Condition-dependent evidence also appears in a model's response to changes in the text condition. SD-MIA studies pre-training membership under black-box access by perturbing the original textual instruction and measuring the relevance of generated images to the queried image~\citep{qi2026sdmia}. The success of its method on a pretrained model provides a useful insight that membership evidence also resides in the local behavior across text conditions.

\begin{figure}[t]
 \centering
 \includegraphics[width=\linewidth]{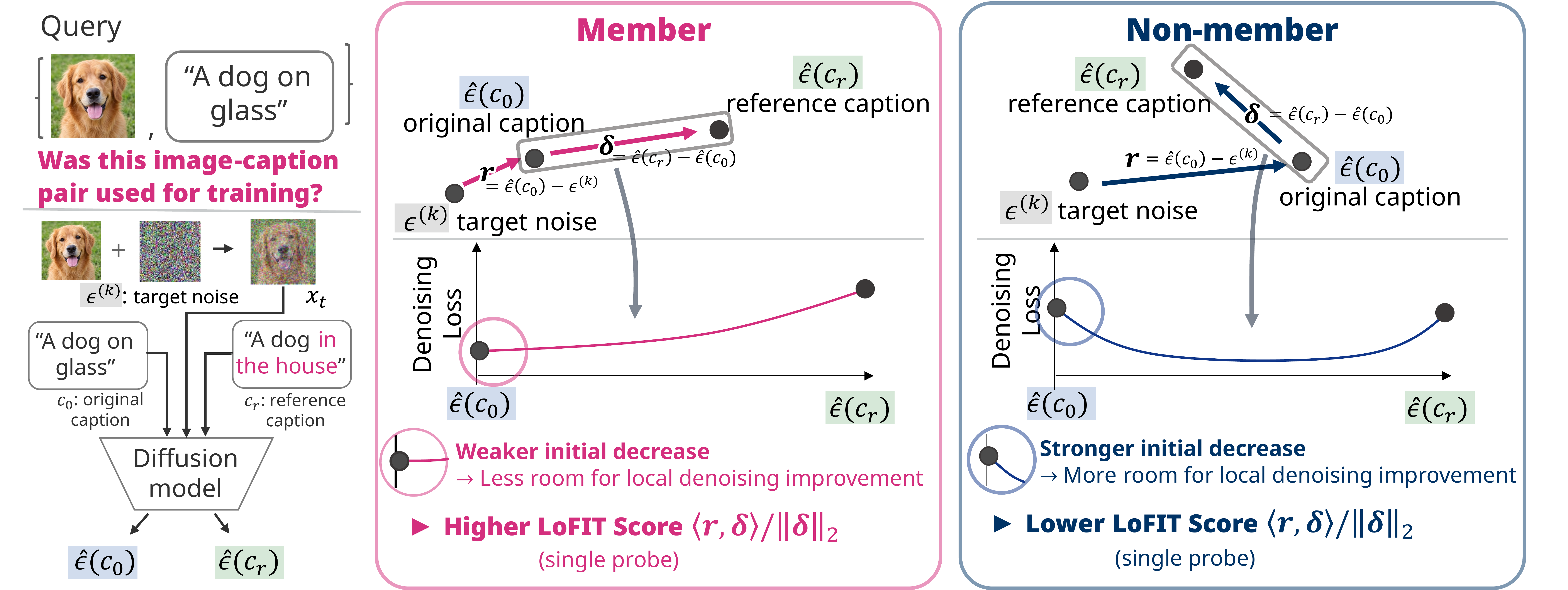}
 \caption{Illustration of LoFIT for a queried image-caption pair. At a fixed noisy input, $r$ denotes the original-caption prediction error relative to the target noise, and $\delta$ the prediction shift toward a reference-caption prediction. LoFIT measures the initial denoising-loss response along this shift through the normalized alignment $\langle r,\delta\rangle/\|\delta\|_2$. Members exhibit a weaker initial loss decrease than non-members, indicating less room for local denoising improvement and yielding a higher score. The curves are conceptual; measured class distributions appear in Figure~\ref{fig:landscape-parabola-minima}.}
 \label{fig:alignment-concept}
\end{figure}

CLiD and SD-MIA provide complementary motivations. CLiD shows that membership information can lie in the image-text relationship, while SD-MIA shows that changing the text condition can expose it. These findings motivate us to ask how a change in text condition affects the denoising fit for the same image. With access to the noise predictor, we examine this question by comparing predictions across conditions while holding the noisy input and target noise fixed.

Since training minimizes denoising loss on observed image-caption pairs, we hypothesize that predictions for members have less potential for further reduction in denoising loss through conditional adjustments than those for non-members. Based on this hypothesis, we introduce \emph{LoFIT} (Local Conditional Fit), a membership score that measures the local denoising-loss response to conditional adjustments. LoFIT uses predictions under alternative captions to probe whether the original prediction can be locally improved. For each alternative, it measures the initial change in denoising loss per unit movement toward that prediction, then averages these responses across reference captions and diffusion states. Higher scores indicate less scope for local improvement on average and provide evidence of membership. LoFIT computes this score using only forward predictions, without access to model gradients.

We evaluate LoFIT on four datasets spanning fine-tuned and pretrained diffusion models. It outperforms the evaluated baselines, including CLiD~\citep{zhai2024clid} and PFAMI~\citep{fu2023pfami}, in AUROC, reaching $0.935$, $0.954$, and $0.896$ on fine-tuned SD v1.4 and $0.602$ on pretrained SD v1.5 (Section~\ref{sec:main-results}). Across training checkpoints, its discrimination strengthens as fixed image-caption pairs are used for training, while absolute denoising loss remains a weak discriminator (Section~\ref{sec:training-analysis}). On identical predictions, controlled decompositions locate membership separation in the first-order alignment response (Section~\ref{sec:signed-response}). Matched-budget comparisons show LoFIT's advantage at equal query cost (Section~\ref{sec:budget}).

Our contributions are:
\begin{itemize}
 \item We identify a membership signal in the local denoising-loss response to conditional prediction adjustments: members exhibit less potential for loss reduction than non-members.
 \item We introduce \emph{LoFIT}, which measures this response per unit prediction change and aggregates it across reference captions and diffusion states using only forward queries.
 \item We evaluate LoFIT on four evaluation sets and use controlled comparisons to identify the contribution of its first-order response.
\end{itemize}

\section{Background and Related Work}
\label{sec:related}

\subsection{Related work}
\label{sec:other-related}

\paragraph{Denoising-based membership inference.}
SecMI uses deterministic reconstruction errors at intermediate steps~\citep{duan2023secmi}. PIA uses the noise predicted at $t=0$ to perturb the image, then measures how accurately the denoiser recovers that noise at a later timestep~\citep{kong2024pia}. SimA uses the norm of the predicted score as its membership statistic~\citep{rao2025sima}. Quantile regression learns image-specific reconstruction-loss thresholds from non-member data and identifies members whose loss falls below the predicted threshold~\citep{tang2024quantile}. Our method also uses denoising predictions, but probes how their error changes along normalized prediction shifts on a shared noisy input.

\paragraph{Conditional comparisons.}
Text conditioning influences training-image replication~\citep{somepalli2023mitigating}, and the magnitude of text-conditional predictions can identify memorized prompts~\citep{wen2024memorization}. CLiD compares denoising losses under the original caption and alternative conditions to probe the learned image-caption relationship~\citep{zhai2024clid}. MoFit addresses missing training captions by optimizing a surrogate image and conditioning embedding, then measuring conditional versus unconditional loss discrepancy~\citep{jeon2026mofit}. Its gradient-based optimization requires stronger model access than our forward-only interface. We use the supplied caption and probe local conditional fit without such optimization. Appendix~\ref{sec:alignment-analysis} relates our score to conditional loss gaps.

\paragraph{Generated-image signals.}
Black-box attacks infer membership from generated images without accessing model weights or intermediate denoising predictions~\citep{li2025blackbox}. Noise as a Probe injects image semantics into the initial noise~\citep{lian2026noise}. SD-MIA perturbs textual instructions and measures how the generated images remain related to the target image~\citep{qi2026sdmia}.

\subsection{From conditional membership signals to local conditional fit}
\label{sec:conditional-signals}
\label{sec:background-clid}
\label{sec:background-sdmia}

We examine CLiD and SD-MIA more closely because they connect membership evidence to two aspects of conditioning: how well the original caption fits the image and how model behavior changes when the condition changes.

\paragraph{CLiD: conditional likelihood discrepancy.}
CLiD~\citep{zhai2024clid} asks how much the supplied caption increases the likelihood of an image:
\begin{equation}
 I_\theta(x_0,c_0)=\log p_\theta(x_0|c_0)-\log p_\theta(x_0).
 \label{eq:clid-likelihood}
\end{equation}
This is the pointwise mutual information under the model's joint distribution; its expectation gives mutual information. Under CLiD's conditional-overfitting assumption, this discrepancy provides membership evidence beyond the image's marginal likelihood.

CLiD approximates this quantity through the diffusion variational bound, using denoising-loss differences. It also compares against reduced captions $c$ to mitigate estimation variability:
\begin{equation}
 \mathcal D_{\mathrm{CLiD}}(x_0,c_0;c)
 =\mathbb E_{t,\epsilon}\!\left[
 \|\epsilon_\theta(x_t,t,c)-\epsilon\|_2^2
 -\|\epsilon_\theta(x_t,t,c_0)-\epsilon\|_2^2
 \right].
 \label{eq:clid-gap}
\end{equation}
Here $x_t$ is the noisy image at timestep $t$, and $\epsilon\sim\mathcal N(0,I)$ is the injected noise; both predictions use the same $x_t$, $t$, and $\epsilon$. The empty caption supplies the unconditional comparison, while other reduced conditions retain partial caption information. A larger gap means that the original caption improves denoising more than the reduced condition. CLiD combines these gaps with the original-caption loss, using either a thresholded scalar score or a learned classifier. Its central contribution is to exploit the learned image-caption association as membership evidence, beyond absolute denoising accuracy.

\paragraph{SD-MIA: stability under textual perturbations.}
SD-MIA~\citep{qi2026sdmia} studies whether the image-caption association persists under textual perturbations. It argues that locally overfitted text-to-image mappings can make generations for members less sensitive to such changes. For a caption embedding $e_0$ and a small perturbation $h$, its motivating sensitivity is
\begin{equation}
 \bigl|p_\theta(x_0|e_0+h)-p_\theta(x_0|e_0)\bigr|
 \approx\bigl|\langle\nabla_e p_\theta(x_0|e_0),h\rangle\bigr| \approx 0.
 \label{eq:sdmia-sensitivity}
\end{equation}
This probability and its gradient are unavailable under SD-MIA's generation-only access. Instead, it constructs coherent caption variants through token, style, and semantic edits, and contrasts the CLIP~\citep{radford2021clip} relevance of their generated images to the target with that under the original caption. Its signal therefore concerns how well generated content remains associated with the target when conditioning changes, without observing denoising losses or target-model gradients.

\paragraph{Local conditional fit hypothesis.}
Together, conditional overfitting and stability under textual perturbations motivate our hypothesis: member captions are closer to a denoising optimum relative to alternative conditions. Training directly minimizes denoising loss on observed image-caption pairs, suggesting less potential for local denoising improvement for members. Section~\ref{sec:method} measures this response along alternative-caption prediction shifts while holding the noisy input and target noise fixed.

\section{Measuring Local Conditional Fit}
\label{sec:method}

We assess whether member captions are closer to a denoising optimum by measuring the local loss response toward alternative-caption predictions. Denoising loss is the training objective and is connected to likelihood through the diffusion variational bound~\citep{ho2020ddpm}.

For a queried pair $(x_0,c_0)$, let $\mathcal C=\operatorname{supp}p(c)$ be the support of the condition prior and $\mathcal U(c_0)\subseteq\mathcal C$ a set of comparison captions containing $c_0$. At each timestep $t$ and noise draw $\epsilon\sim\mathcal N(0,I)$, all captions use the same noisy input $x_t$ and target noise $\epsilon$. The denoising loss and optimality criterion are
\begin{align}
 \mathcal L(x_0,c)&=\mathbb E_{t,\epsilon}
 \bigl[\|\epsilon_\theta(x_t,t,c)-\epsilon\|_2^2\bigr],\nonumber\\
 \mathcal L(x_0,c_0)&\leq\mathcal L(x_0,c)
 \quad\text{for all }c\in\mathcal U(c_0).
 \label{eq:conditional-local-objective}
\end{align}
This criterion concerns optimality over the comparison set. Locality in LoFIT refers to infinitesimal movement in prediction space. We measure the response at each noisy state, then aggregate across states and references.

\subsection{A first-order criterion for conditional fit}
\label{sec:first-order-fit}

Squared error is convex in prediction space. For a given noisy image $x_t$, timestep $t$, and noise $\epsilon$, the following first-order condition is sufficient for $c_0$ to minimize the denoising loss over $\mathcal U(c_0)$~\citep[Section~4.2.3]{boyd2004convex}:
\begin{align}
 &\bigl\langle\epsilon_\theta(x_t,t,c_0)-\epsilon,\,
 \epsilon_\theta(x_t,t,c)-\epsilon_\theta(x_t,t,c_0)\bigr\rangle\geq0,\nonumber\\
 &\qquad\text{for all }c\in\mathcal U(c_0),\nonumber\\
 &\Longrightarrow\quad
 c_0\in\arg\min_{c\in\mathcal U(c_0)}
 \|\epsilon_\theta(x_t,t,c)-\epsilon\|_2^2.
 \label{eq:conditional-optimality}
\end{align}
The inner product aligns prediction error with the conditional prediction shift. Nonnegative alignment for every comparison caption ensures that none yields a lower loss. If this holds at every noisy state, averaging gives Equation~\ref{eq:conditional-local-objective}. This criterion motivates our alignment score.

\subsection{Measuring alignment}
\label{sec:posterior}
\label{sec:scores}

Checking the condition over all captions is impractical. LoFIT instead evaluates the inner product using a finite set of reference captions from $\mathcal U(c_0)$. For a reference caption $c$, the original prediction error $r$ and the conditional prediction shift $\delta$ are
\begin{align}
 r&=\epsilon_\theta(x_t,t,c_0)-\epsilon,\nonumber\\
 \delta&=\epsilon_\theta(x_t,t,c)-\epsilon_\theta(x_t,t,c_0).
 \label{eq:residual-shift}
\end{align}
For $\delta\ne0$, we use the \emph{normalized prediction shift} $\delta/\|\delta\|_2$ to compare directions independently of shift magnitude. The \emph{alignment response} is
\begin{equation}
 v=\frac{\langle r,\delta\rangle}{\|\delta\|_2}.
 \label{eq:alignment}
\end{equation}
Normalization preserves the sign of the inner product. The normalized response can also be interpreted as half the directional derivative of the denoising loss along the normalized prediction shift $\delta/\|\delta\|_2$. For a step of length $s$ toward the reference prediction,
\begin{equation}
 \left.\frac{\mathrm d}{\mathrm d s}
 \left\|r+s\frac{\delta}{\|\delta\|_2}\right\|_2^2
 \right|_{s=0}
 =2\frac{\langle r,\delta\rangle}{\|\delta\|_2}=2v.
 \label{eq:alignment-derivative}
\end{equation}
Thus $v<0$ indicates an initial loss decrease toward the reference prediction, while $v\geq0$ indicates no first-order decrease. At the reference prediction, $s=\|\delta\|_2$; the finite-step loss change $2sv+s^2$ can therefore be positive even when $v<0$. CLiD's discrepancy term uses this finite loss gap, whereas LoFIT isolates the initial response (Section~\ref{sec:signed-response}). LoFIT's response is local in prediction space even when the two captions differ substantially.

Score composition gives this local response a conditional interpretation. Compositional diffusion and classifier-free guidance combine predictions under different conditions to control generation~\citep{liu2022composable,ho2022cfg}. At a fixed noise level, a convex combination of exact conditional scores is itself the score of a normalized geometric mixture of their densities. Noise predictions estimate these scores up to a shared scale, so interpolating predictions forms composed score estimates. LoFIT tests whether a small contribution from an alternative condition initially reduces the original denoising error, measuring local fit within this composed family.

\subsection{Aggregating alignment responses}
\label{sec:aggregation}

To compute LoFIT, let $\mathcal R$ index reference captions $c_j$, including caption edits, randomly sampled captions, and the empty prompt (Appendix~\ref{sec:candidates}). We evaluate $T$ timesteps with $K$ Gaussian noise draws each, using $i=(t,k)$ to index the $TK$ states. For each draw $\epsilon^{(k)}\sim\mathcal N(0,I)$, we form
\begin{equation}
 x_{t,k}=\sqrt{\bar\alpha_t}\,x_0+\sqrt{1-\bar\alpha_t}\,\epsilon^{(k)}.
 \label{eq:noisy-input}
\end{equation}
Here $\bar\alpha_t$ is the diffusion noise schedule, and $x_0$ is the queried image or its latent representation. Each comparison uses the same noisy image $x_{t,k}$ and target noise $\epsilon^{(k)}$ for both captions. At state $i$, $r_i$ denotes the original prediction error, while $\delta_{ij}$ and $v_{ij}$ denote the prediction shift and alignment for reference $j$. Let $\mathcal A$ be the set of valid state-reference pairs $(i,j)$ after omitting zero prediction shifts; Appendix~\ref{sec:zero-shifts} specifies the numerical tolerance. The LoFIT score is
\begin{equation}
 S(x_0,c_0)=
 \begin{cases}
 |\mathcal A|^{-1}\displaystyle\sum_{(i,j)\in\mathcal A}v_{ij}, & |\mathcal A|>0,\\
 0, & |\mathcal A|=0.
 \end{cases}
 \label{eq:unified}
\end{equation}
When every pair is valid, $|\mathcal A|=TK|\mathcal R|$. Higher $S$ indicates less initial denoising improvement and provides membership evidence. The query cost is $TK(1+|\mathcal R|)$, reusing the original prediction at each state. Appendices~\ref{sec:minimum-analysis} and~\ref{sec:signed-details} analyze loss minima and the contribution of the response sign.

\section{Experiments}
\label{sec:experiments}

We evaluate LoFIT on the full evaluation sets and compare it with published attacks. Controlled readout comparisons use the same reference predictions and diffusion states.

\begin{table}[!t]
\centering
\caption{Full-set membership inference on fine-tuned SD v1.4, SD v2.1, and SD-XL. Bold marks the best value per backbone and metric; dashes denote unreported results among the listed methods.}
\label{tab:main-summary}
\fontsize{8}{10}\selectfont
\renewcommand{\arraystretch}{0.9}
\setlength{\tabcolsep}{3pt}
\begin{tabular*}{\linewidth}{@{\extracolsep{\fill}}lcccccc@{}}
\toprule
& \multicolumn{2}{c}{Flickr30k} & \multicolumn{2}{c}{COCO} & \multicolumn{2}{c}{Pokemon} \\
\cmidrule(lr){2-3}\cmidrule(lr){4-5}\cmidrule(lr){6-7}
Method & AUROC & TPR@1\%FPR & AUROC & TPR@1\%FPR & AUROC & TPR@1\%FPR \\
\midrule
\multicolumn{7}{@{}l}{\textbf{SD v1.4}} \\
\midrule
Naive loss & 0.524 & 0.013 & 0.527 & 0.018 & 0.576 & 0.029 \\
PIA & 0.507 & 0.010 & 0.503 & 0.012 & 0.552 & 0.027 \\
SecMI & 0.551 & 0.017 & 0.562 & 0.012 & 0.538 & 0.022 \\
SD-MIA & 0.519 & 0.013 & 0.517 & 0.013 & 0.502 & 0.019 \\
PFAMI & 0.659 & 0.052 & 0.715 & 0.073 & 0.685 & 0.031 \\
CLiD$_{\mathrm{th}}$ & 0.718 & 0.063 & 0.745 & 0.072 & 0.695 & 0.058 \\
CLiD$_{\mathrm{vec}}$ & 0.718 & 0.064 & 0.751 & 0.073 & 0.642 & 0.043 \\
\textbf{LoFIT (ours)} & \textbf{0.935} & \textbf{0.454} & \textbf{0.954} & \textbf{0.478} & \textbf{0.896} & \textbf{0.272} \\
\midrule
\multicolumn{7}{@{}l}{\textbf{SD v2.1}} \\
\midrule
Naive loss & 0.519 & 0.012 & 0.516 & 0.014 & 0.557 & 0.019 \\
PIA & 0.505 & 0.011 & 0.500 & 0.008 & 0.546 & 0.029 \\
SecMI & 0.537 & 0.013 & 0.530 & 0.010 & 0.503 & 0.017 \\
PFAMI & 0.631 & 0.038 & 0.653 & 0.060 & 0.631 & 0.029 \\
CLiD$_{\mathrm{th}}$ & 0.705 & 0.088 & 0.681 & 0.070 & 0.686 & 0.026 \\
CLiD$_{\mathrm{vec}}$ & 0.783 & 0.131 & 0.739 & 0.079 & 0.684 & 0.022 \\
\textbf{LoFIT (ours)} & \textbf{0.894} & \textbf{0.296} & \textbf{0.851} & \textbf{0.166} & \textbf{0.812} & \textbf{0.130} \\
\midrule
\multicolumn{7}{@{}l}{\textbf{SD-XL}} \\
\midrule
Naive loss & 0.514 & 0.011 & 0.516 & 0.010 & 0.530 & 0.014 \\
PIA & 0.498 & 0.009 & 0.501 & 0.008 & 0.517 & 0.012 \\
SecMI & 0.515 & 0.011 & 0.506 & 0.012 & 0.477 & 0.019 \\
PFAMI & 0.660 & 0.069 & 0.657 & 0.052 & 0.664 & 0.048 \\
CLiD$_{\mathrm{th}}$ & 0.866 & 0.188 & 0.819 & 0.114 & 0.804 & 0.127 \\
CLiD$_{\mathrm{vec}}$ & 0.880 & 0.217 & 0.836 & 0.121 & 0.784 & 0.084 \\
\textbf{LoFIT (ours)} & \textbf{0.978} & \textbf{0.719} & \textbf{0.969} & \textbf{0.631} & \textbf{0.883} & \textbf{0.351} \\
\bottomrule
\end{tabular*}
\end{table}

\subsection{Settings and comparisons}
\label{sec:settings}

\paragraph{Reference construction and score computation.}
References comprise randomly sampled captions, the empty prompt, and language-model edits. LoFIT computes each score directly from forward predictions using a fixed reference-construction procedure, without fitting an attack classifier. The configurations below specify the reference sets and query allocations.

\paragraph{Models and data.}
We evaluate fine-tuned SD v1.4, SD v2.1-base ($\epsilon$-prediction), and SD-XL base 1.0 on Flickr30k~\citep{young2014flickr30k}, COCO~\citep{lin2014coco}, and Pokemon~\citep{pinkney2022pokemon}, with $10{,}000/10{,}000$, $2{,}500/2{,}500$, and $416/417$ members/non-members, respectively. LAION-MI uses pretrained SD v1.5 with $6{,}400$ examples per class, matched in CLIP~\citep{radford2021clip} image-text similarity using $20$ bins~\citep{dubinski2024laionmi}. Appendix~\ref{sec:protocols} gives the training configurations, and Appendix~\ref{sec:laion-distribution} details LAION-MI matching and preprocessing.

\paragraph{Default evaluation configuration.}\label{sec:default-configuration}
All three fine-tuned backbones use the same diffusion-state grids and noise-draw counts. SD v2.1 and SD-XL use the reduced reference sets in Appendix~\ref{sec:sd21-configuration}. On the three fine-tuned SD v1.4 models, we use all eligible reference families: random captions, the empty prompt, and language-model rewrites. LAION-MI uses six randomly sampled captions and the empty prompt ($|\mathcal R|=7$); adding rewrites weakens discrimination there (Appendix~\ref{sec:alignment-aggregation}). Flickr30k and LAION-MI use $T=12$ timesteps $\{60,120,\ldots,720\}$ and $K=4$ noise draws; COCO and Pokemon use $T=4$ timesteps ($150$, $300$, $450$, $600$) and $K=4$. On Flickr30k/COCO/Pokemon, SD v1.4 requires $2{,}064/752/1{,}088$ denoiser evaluations per image, and SD v2.1 and SD-XL require $528/176/272$; LAION-MI requires $384$. These counts include the original caption.

\paragraph{Metrics.}
We report AUROC and true-positive rates (TPR) at $1\%$ false-positive rate (FPR) for fine-tuned models and $5\%$ FPR for LAION-MI.

\paragraph{Baselines and controlled comparisons.}
We compare with naive loss, PIA~\citep{kong2024pia}, SecMI~\citep{duan2023secmi}, SD-MIA~\citep{qi2026sdmia}, CLiD~\citep{zhai2024clid}, and PFAMI~\citep{fu2023pfami}. CLiD uses character-based caption thirds and the empty prompt at three fixed timesteps. PFAMI measures relative loss fluctuations under image crops, with calibration by the corresponding pretrained backbone on fine-tuned targets and no calibration on LAION-MI. Appendix~\ref{sec:baseline-protocols} details these settings and deviations from the original protocols. To isolate the contribution of LoFIT's scoring rule, we also evaluate CLiD variants using our expanded reference captions, including LLM rewrites and random captions, or our timestep configuration. Table~\ref{tab:directional-readout} compares readouts on identical predictions, while Section~\ref{sec:budget} examines query budgets.

\Needspace{5\baselineskip}
\subsection{Results}
\label{sec:main-results}
\label{sec:budget}

\begin{table}[!htb]
\centering\small
\caption{Membership inference on pretrained SD v1.5 (LAION-MI). Bold marks the best values; underlining marks the second-best TPR.}
\label{tab:laion-summary}
\setlength{\tabcolsep}{2pt}
\begin{tabular*}{\linewidth}{@{\extracolsep{\fill}}lcccccccc@{}}
\toprule
Metric & Naive loss & PIA & SecMI & SD-MIA & PFAMI & CLiD$_{\mathrm{th}}$ & CLiD$_{\mathrm{vec}}$ & \textbf{LoFIT} \\
\midrule
AUROC & 0.527 & 0.521 & 0.512 & 0.571 & 0.516 & 0.557 & 0.548 & \textbf{0.602} \\
TPR@5\%FPR & 0.072 & 0.072 & 0.063 & \textbf{0.112} & 0.066 & 0.075 & 0.076 & \underline{0.101} \\
\bottomrule
\end{tabular*}
\end{table}

\paragraph{Membership inference performance.}
LoFIT outperforms the evaluated baselines in both AUROC and low-FPR TPR on all three datasets with fine-tuned SD v1.4, SD v2.1, and SD-XL (Table~\ref{tab:main-summary}). On pretrained SD v1.5, LoFIT achieves the highest evaluated AUROC of $0.602$ on LAION-MI, with TPR $0.101$ at $5\%$ FPR (Table~\ref{tab:laion-summary}). Appendix~\ref{sec:video-results} reports additional video membership inference results on CogVideoX-2b and Wan2.1.

\paragraph{Query efficiency.}

\begin{figure}[!htb]
    \centering
    \includegraphics[width=.97\linewidth]{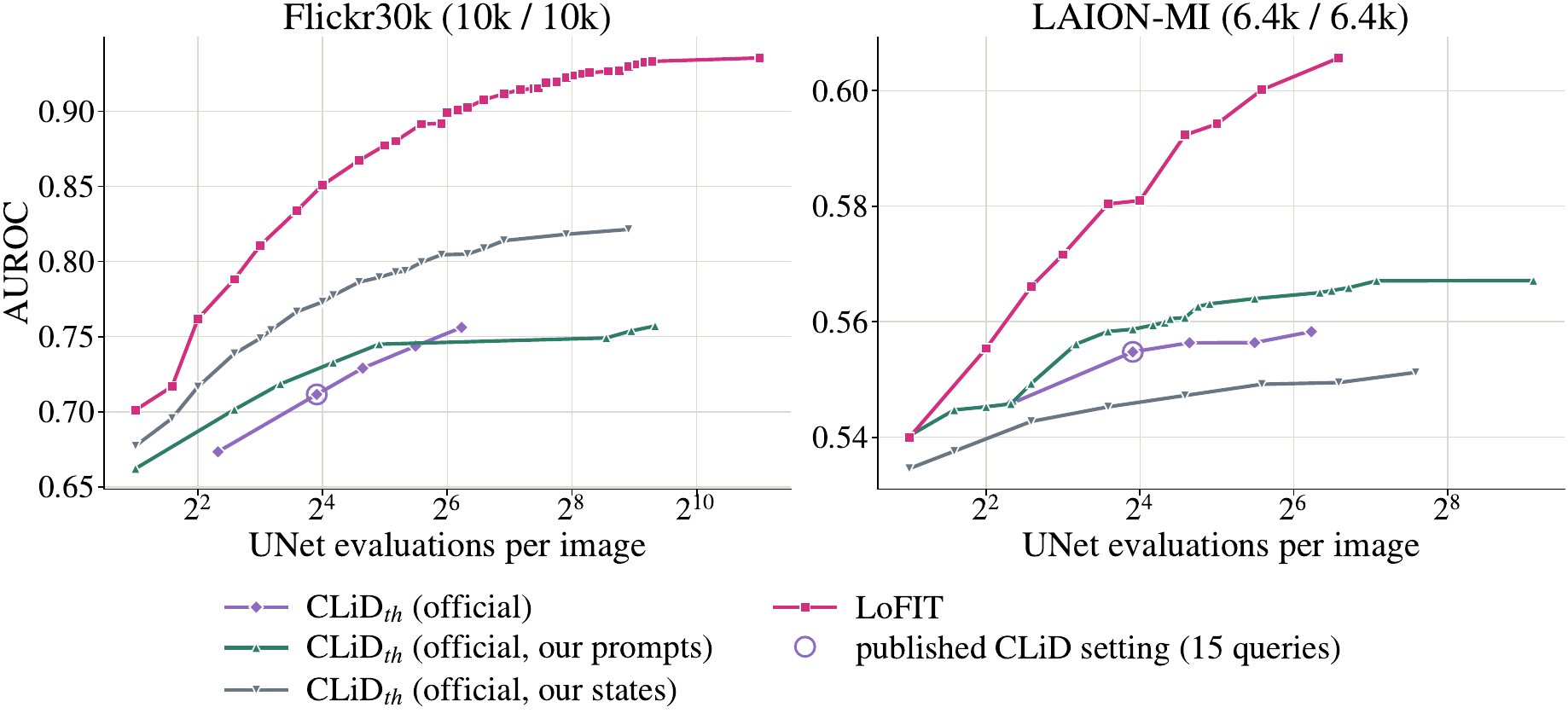}
    \caption{Query efficiency on Flickr30k and LAION-MI. CLiD curves use the released statistic with two-fold cross-fitted calibration (Appendix~\ref{sec:baseline-protocols}); ``our prompts'' and ``our states'' denote our reference captions and timestep grid. Open circles mark the original $15$-query configuration.}
    \label{fig:budget}
\end{figure}

LoFIT outperforms CLiD$_{\mathrm{th}}$ at matched query budgets, including variants using our captions or timestep grid (Figure~\ref{fig:budget}). At $15$ denoiser evaluations, AUROC is $0.834$ versus $0.767$ for the strongest CLiD variant on Flickr30k, and $0.580$ versus $0.559$ on LAION-MI. Appendix~\ref{sec:baseline-protocols} specifies calibration.

On identical reference predictions and diffusion states, LoFIT outperforms CLiD$_{\mathrm{th}}$ with both fixed $\alpha=0.5$ and oracle $\alpha$ (Table~\ref{tab:reference-state-readout}). With all reference families, its AUROC advantage over oracle CLiD$_{\mathrm{th}}$ is $0.081$--$0.130$ on the fine-tuned models at the same query cost. Appendix~\ref{sec:clid-matched-controls} specifies scaling and oracle selection.

With the empty reference alone, LoFIT also outperforms both CLiD$_{\mathrm{th}}$ variants across all four datasets. On Flickr30k and COCO, empty-only LoFIT exceeds even oracle CLiD$_{\mathrm{th}}$ using all reference families.

\begin{table}[!htb]
\centering\small
\caption{LoFIT and CLiD$_{\mathrm{th}}$ on identical reference predictions and diffusion states. Bold marks the best value in each reference group.}
\label{tab:reference-state-readout}
\fontsize{8}{10}\selectfont
\setlength{\tabcolsep}{1.5pt}
\begin{tabular*}{\linewidth}{@{\extracolsep{\fill}}lcccccccc@{}}
\toprule
& \multicolumn{2}{c}{Flickr30k} & \multicolumn{2}{c}{COCO} & \multicolumn{2}{c}{Pokemon} & \multicolumn{2}{c}{LAION-MI} \\
\cmidrule(lr){2-3}\cmidrule(lr){4-5}\cmidrule(lr){6-7}\cmidrule(lr){8-9}
Method & AUROC & \shortstack{TPR@1\%\\FPR} & AUROC & \shortstack{TPR@1\%\\FPR} & AUROC & \shortstack{TPR@1\%\\FPR} & AUROC & \shortstack{TPR@5\%\\FPR} \\
\midrule
\multicolumn{9}{@{}l}{\emph{All reference families (LAION-MI: 36 references; Table~\ref{tab:laion-summary} uses 7)}} \\
\textbf{LoFIT} & \textbf{0.935} & \textbf{0.454} & \textbf{0.954} & \textbf{0.478} & \textbf{0.896} & \textbf{0.272} & \textbf{0.573} & \textbf{0.085} \\
CLiD$_{\mathrm{th}}$ ($\alpha=0.5$) & 0.731 & 0.106 & 0.752 & 0.130 & 0.764 & 0.058 & 0.563 & 0.084 \\
CLiD$_{\mathrm{th}}$ (oracle $\alpha$) & 0.813 & 0.092 & 0.824 & 0.083 & 0.815 & 0.079 & 0.569 & 0.083 \\
\midrule
\multicolumn{9}{@{}l}{\emph{Empty reference only}} \\
\textbf{LoFIT} & \textbf{0.900} & \textbf{0.268} & \textbf{0.914} & \textbf{0.287} & \textbf{0.781} & \textbf{0.070} & \textbf{0.606} & \textbf{0.094} \\
CLiD$_{\mathrm{th}}$ ($\alpha=0.5$) & 0.730 & 0.112 & 0.736 & 0.094 & 0.720 & 0.053 & 0.549 & 0.089 \\
CLiD$_{\mathrm{th}}$ (oracle $\alpha$) & 0.806 & 0.146 & 0.799 & 0.097 & 0.756 & 0.053 & 0.553 & 0.081 \\
\bottomrule
\end{tabular*}
\end{table}

\paragraph{Image membership with generated captions.}
With BLIP-2 OPT-6.7b~\citep{li2023blip2} captions, prompt-based LoFIT achieves higher AUROC than official CLiD$_{\mathrm{th}}$ with oracle $\alpha$. Using random captions and the empty reference, LoFIT reaches $0.641$, $0.669$, and $0.621$ AUROC on fine-tuned SD v1.4 for Flickr30k, COCO, and Pokemon, respectively, compared with $0.603$, $0.650$, and $0.576$ for CLiD$_{\mathrm{th}}$. On LAION-MI, the corresponding AUROC values are $0.546$ and $0.535$. Gaussian conditioning probes provide further improvements on the fine-tuned models. Appendix~\ref{sec:generated-captions} reports the setup and full comparisons.

\section{Understanding the Membership Signal}
\label{sec:analysis}

The following analyses use the SD v1.4 fine-tunes and pretrained SD v1.5.

\subsection{Early membership detection during fine-tuning}
\label{sec:training-analysis}

\begin{figure}[!htb]
 \centering
 \includegraphics[width=.86\linewidth,trim=0 0 0 24.2,clip]{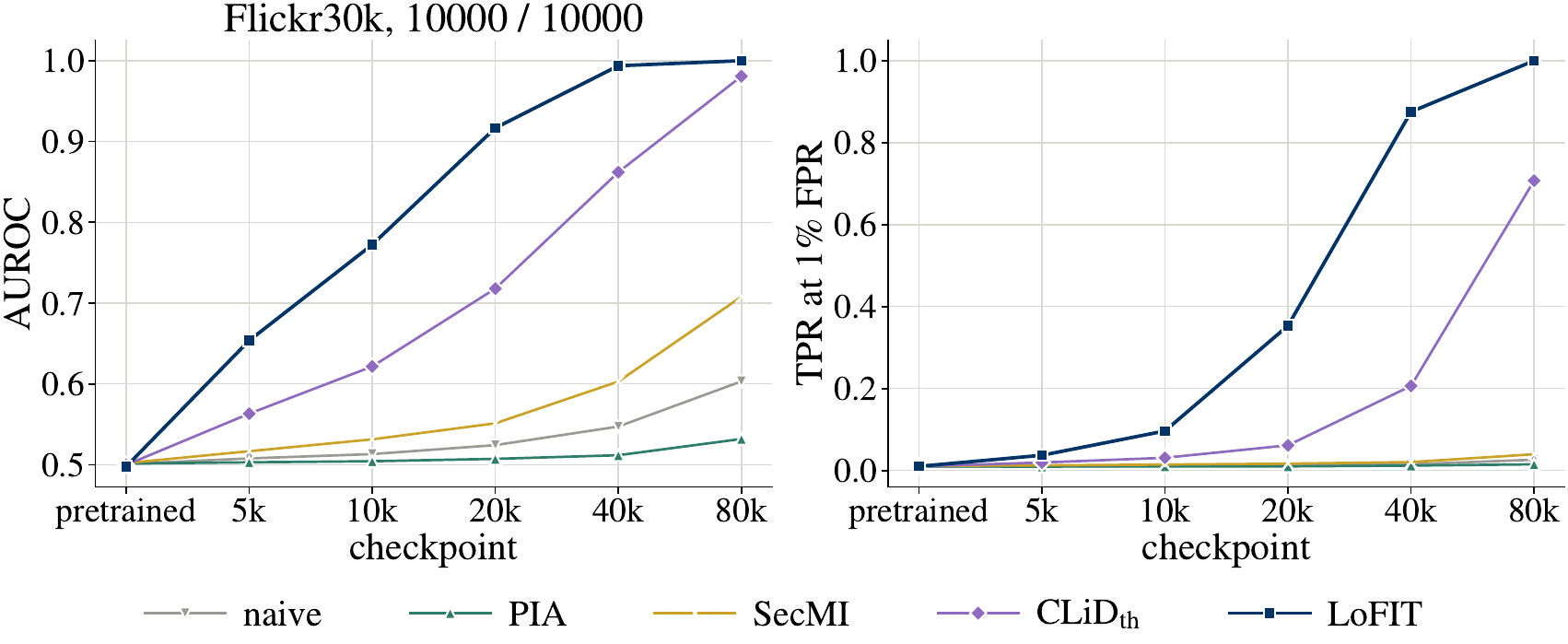}
 \caption{Membership inference across Flickr30k fine-tuning checkpoints, including official CLiD$_{\mathrm{th}}$. Membership denotes participation in this fine-tuning run.}
 \label{fig:alignment-training}
\end{figure}

LoFIT detects membership earlier and approaches saturation sooner than official CLiD$_{\mathrm{th}}$ (Figure~\ref{fig:alignment-training}). At $20$k steps, LoFIT achieves AUROC $0.917$ and TPR $0.353$ at $1\%$ FPR. By $40$k, these reach $0.994$ and $0.875$, while CLiD$_{\mathrm{th}}$ continues improving through $80$k steps. Absolute denoising loss remains a weak discriminator across these checkpoints. We evaluate the same $10{,}000$ member and $10{,}000$ non-member Flickr30k pairs throughout. LoFIT keeps its references and $12\times4$ diffusion states fixed; CLiD$_{\mathrm{th}}$ uses the released caption reductions and three-state configuration.

\subsection{Caption and Gaussian conditioning probes}
\label{sec:reference-directions}

Table~\ref{tab:reference-directions} compares caption and Gaussian probes at equal single-reference cost ($2TK$ denoiser evaluations). Both use the same images, diffusion-state grids, and alignment score; original-caption losses match across all datasets (Appendix~\ref{sec:random-directions}). AUROC is averaged across references within each group.

\begin{table}[H]
\centering\small
\caption{Single-reference AUROC for combined caption edits and Gaussian probes.}
\label{tab:reference-directions}
\setlength{\tabcolsep}{6pt}
\begin{tabular}{@{}lcccc@{}}
\toprule
Reference directions & Flickr30k & COCO & Pokemon & LAION-MI \\
\midrule
Combined edits & 0.880 & 0.909 & 0.810 & 0.569 \\
Gaussian, $h=160$ & 0.854 & 0.903 & 0.805 & 0.534 \\
\bottomrule
\end{tabular}
\end{table}

Gaussian probes approach combined edits on COCO and Pokemon. On LAION-MI, the empty caption achieves AUROC $0.606$, compared with $0.534$ for Gaussian probes at $h=160$. Pooling caption and Gaussian responses further improves discrimination on the fine-tuned models (Appendix~\ref{sec:random-directions}). Gaussian probes require direct modification of conditioning embeddings, extending access beyond LoFIT's text-condition gray-box interface.

\begin{figure}[!t]
 \centering
 \includegraphics[width=.90\linewidth]{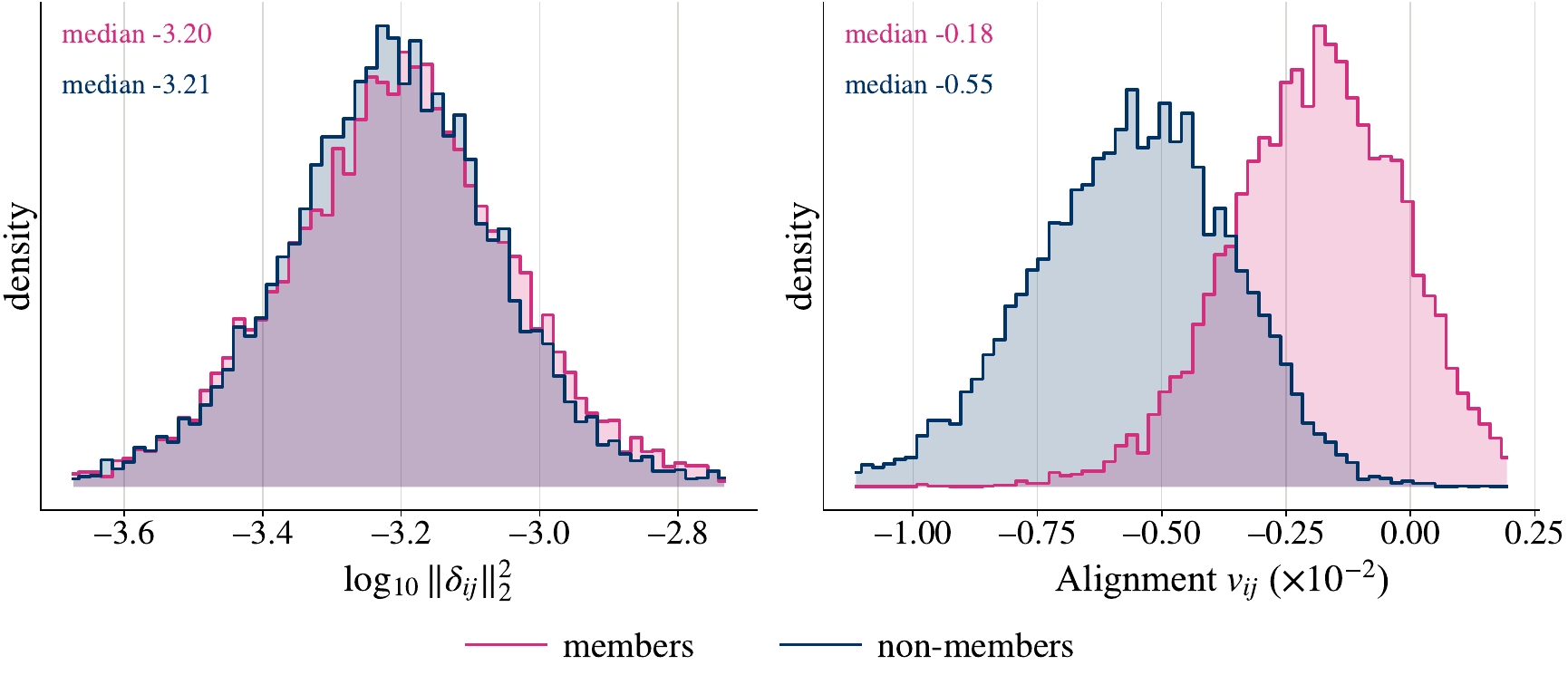}
 \caption{Conditional prediction statistics on Flickr30k. Left: distributions of squared shift magnitude $\|\delta_{ij}\|_2^2$ on a logarithmic scale. Right: normalized residual interaction $\langle r_i,\delta_{ij}\rangle/\|\delta_{ij}\|_2$, the LoFIT response. Shift magnitudes largely overlap, while the first-order response separates the classes. Magenta denotes members and blue denotes non-members. Full profiles appear in Appendix~\ref{sec:residual-interaction}.}
 \label{fig:shift-profile-summary}
\end{figure}

\subsection{Alignment isolates the directional membership signal}
\label{sec:signed-response}

The discrepancy term in CLiD measures conditional denoising-loss gaps~\citep{zhai2024clid}, while LoFIT isolates their first-order response. For a fixed noisy image and target noise, the gap decomposes as
\begin{align}
 \Delta_{ij}&=\|r_i+\delta_{ij}\|_2^2-\|r_i\|_2^2
 =2\langle r_i,\delta_{ij}\rangle+\|\delta_{ij}\|_2^2,\label{eq:gap}\\
 \frac{\Delta_{ij}}{\|\delta_{ij}\|_2}
 &=\underbrace{2v_{ij}}_{\text{first-order response}}
 +\underbrace{\|\delta_{ij}\|_2}_{\text{shift magnitude}}.
 \label{eq:normalized-gap}
\end{align}
The normalized gap adds shift magnitude to LoFIT's first-order response $2v_{ij}$. Shift magnitudes largely overlap between the classes, whereas the response is more negative for non-members (Figure~\ref{fig:shift-profile-summary}; full profiles in Appendix~\ref{sec:residual-interaction}).

On identical predictions under the default configuration, LoFIT outperforms both raw and normalized gaps in AUROC and low-FPR TPR across all four datasets (Table~\ref{tab:directional-readout}). Isolating the first-order response therefore improves discrimination beyond normalization alone.

\begin{table}[!htb]
\centering\small
\caption{Readout comparison on identical predictions and diffusion states. Bold marks the best value per metric.}
\label{tab:directional-readout}
\setlength{\tabcolsep}{2pt}
\begin{tabular*}{\linewidth}{@{\extracolsep{\fill}}lcccccccc@{}}
\toprule
& \multicolumn{2}{c}{Flickr30k} & \multicolumn{2}{c}{COCO} & \multicolumn{2}{c}{Pokemon} & \multicolumn{2}{c}{LAION-MI} \\
\cmidrule(lr){2-3}\cmidrule(lr){4-5}\cmidrule(lr){6-7}\cmidrule(lr){8-9}
Readout & AUROC & \shortstack{TPR@1\%\\FPR} & AUROC & \shortstack{TPR@1\%\\FPR} & AUROC & \shortstack{TPR@1\%\\FPR} & AUROC & \shortstack{TPR@5\%\\FPR} \\
\midrule
Loss gap $\Delta_{ij}$ & 0.813 & 0.092 & 0.818 & 0.078 & 0.804 & 0.065 & 0.549 & 0.076 \\
Normalized gap & 0.906 & 0.334 & 0.930 & 0.380 & 0.882 & 0.127 & 0.567 & 0.085 \\
Shift magnitude $\|\delta_{ij}\|_2$ & 0.503 & 0.012 & 0.487 & 0.014 & 0.457 & 0.024 & 0.480 & 0.053 \\
\textbf{LoFIT} & \textbf{0.935} & \textbf{0.454} & \textbf{0.954} & \textbf{0.478} & \textbf{0.896} & \textbf{0.272} & \textbf{0.602} & \textbf{0.101} \\
\bottomrule
\end{tabular*}
\end{table}
\FloatBarrier
\section{Conclusion}
\label{sec:conclusion}

We introduced LoFIT, a membership inference method that measures local conditional fit for a queried image-caption pair. Members exhibit less potential for denoising-loss reduction toward alternative-caption predictions. LoFIT captures this difference through the normalized first-order response and averages it across reference captions and diffusion states using only forward queries.

Experiments across multiple diffusion backbones show improved membership discrimination in both fine-tuning and pretraining settings, with additional results extending LoFIT to video generation. Analysis of training dynamics shows that LoFIT detects membership earlier in fine-tuning than the baselines. LoFIT also improves the trade-off between membership inference performance and query cost, achieving higher AUROC than CLiD with the same number of denoiser evaluations. These findings establish local conditional fit as a useful signal for auditing training-data use.

\paragraph{Limitations.}
LoFIT's effectiveness depends on the available caption: generated captions retain membership information but yield lower discrimination than supplied training captions (Appendix~\ref{sec:generated-captions}). The method requires forward access to the denoiser at specified noisy inputs, timesteps, and conditions, which generation-only APIs do not provide.

\ifdefined\arxivpreprint\else
\FloatBarrier
\Needspace{8\baselineskip}
\subsection*{AI use statement}

We used generative AI tools to assist with experimental design and
manuscript drafting and editing. These tools also assisted with
LaTeX formatting and the preparation of tables and figures.
We used \texttt{gpt-4.1-mini} to generate reference-caption variants,
as described in Appendix~\ref{sec:candidates}.
The authors take responsibility for the final content of this work,
including all AI-assisted text and figures.

\Needspace{8\baselineskip}
\subsection*{Ethics statement}

This work studies privacy leakage through membership inference in text-to-image diffusion models. LoFIT supports audits of training-data use and assessments of privacy exposure. The same capability can reveal information about the inclusion of image-caption pairs in training data. We intend this method for authorized privacy audits and research on privacy protection, with appropriate safeguards for the queried data.

\Needspace{8\baselineskip}
\subsection*{Reproducibility statement}

Section~\ref{sec:method} defines the LoFIT score and its computation. Section~\ref{sec:experiments} and Appendix~\ref{sec:protocols} document target-model training, data splits, preprocessing, diffusion states, reference construction, and baseline calibration. Controlled comparisons specify the reference sets, states, and readouts used in each configuration. Appendices~\ref{sec:alignment-analysis} and~\ref{sec:direction-design} provide additional analyses of the membership signal and reference construction.
\fi

\FloatBarrier
\clearpage
\bibliographystyle{iclr2027_conference}
\bibliography{main}

\clearpage
\appendix
\flushbottom
\setlength{\tablecaptiontopspace}{10pt}
\setlength{\textfloatsep}{12pt minus 2pt}
\setlength{\floatsep}{8pt minus 2pt}
\setlength{\intextsep}{10pt minus 2pt}
\setlength{\figurecaptionspace}{6pt}
\AddToHook{cmd/section/before}{\FloatBarrier}

\section{Experimental Protocols}
\label{sec:protocols}

This appendix documents target-model training, data construction, reference captions, and baseline evaluation.

\paragraph{Training configurations.}
\label{sec:target-training}
We fine-tune the full UNet using the diffusers text-to-image training script, with the VAE and text encoders frozen. The tokenizer and diffusion noise schedule remain fixed. All nine fine-tuned targets use AdamW with learning rate $10^{-5}$, $\beta=(0.9,0.999)$, weight decay $0.01$, and optimizer epsilon $10^{-8}$. The learning rate is constant with no warmup. Training minimizes epsilon-prediction MSE with timesteps sampled uniformly from a $1{,}000$-step DDPM schedule, without SNR weighting, caption dropout, or noise offset. We use FP16 mixed precision, gradient checkpointing, and gradient clipping at norm $1.0$. EMA weights are maintained during training; all attacks evaluate the raw UNet weights.

SD-XL starts from \texttt{stabilityai/stable-diffusion-xl-base-1.0} and follows the shared training recipe at $1024\times1024$ resolution. Table~\ref{tab:target-training} lists the evaluated checkpoints and effective batch sizes. Steps count optimizer updates, and batch sizes include all devices and gradient accumulation. SD v2.1 Flickr30k uses training seed $0$; the SD v1.4 runs and SD v2.1 COCO/Pokemon do not set a training seed. The data split seed is $0$ for every backbone.

\begin{table}[htbp]
\centering\small
\caption{Target-model training runs and evaluated checkpoints.}
\label{tab:target-training}
\setlength{\tabcolsep}{5pt}
\begin{tabular*}{\linewidth}{@{\extracolsep{\fill}}llrr@{}}
\toprule
Model & Dataset & \shortstack{Effective\\batch} & \shortstack{Evaluated\\update} \\
\midrule
SD v1.4 & Flickr30k & 16 & 20,000 \\
 & COCO & 256 & 1,000 \\
 & Pokemon & 256 & 200 \\
\midrule
SD v2.1 & Flickr30k & 16 & 20,000 \\
 & COCO & 256 & 1,000 \\
 & Pokemon & 256 & 200 \\
\midrule
SD-XL & Flickr30k & 16 & 20,000 \\
 & COCO & 256 & 1,000 \\
 & Pokemon & 256 & 200 \\
\bottomrule
\end{tabular*}
\end{table}

\paragraph{Image splits and captions.}
\label{sec:finetuning-splits}
We shuffle each dataset once with seed $0$, assign the first $n$ images to members, and take the next $n$ as non-members, with $n=10{,}000$ for Flickr30k and $n=2{,}500$ for COCO. Pokemon uses the first $416$ images as members and the remaining $417$ as non-members. Each source row contains one image; all captions associated with a row remain on the same side of the split. We retain only the first Flickr30k caption before shuffling. Every image has one fixed caption throughout training, and the same string serves as its original caption during evaluation. The caption columns are \texttt{caption}, \texttt{captions}, and \texttt{text} for Flickr30k, COCO, and Pokemon, respectively.

\paragraph{Preprocessing.}
We resize the shorter image side to $1024$ pixels for SD-XL and $512$ for SD v1.4, SD v1.5, and SD v2.1, then apply a square center crop at that resolution. Training adds horizontal flipping with probability $0.5$ and normalizes pixels to $[-1,1]$; attack evaluation omits flipping. LoFIT, naive loss, PIA, and SecMI use the VAE posterior mean. The official CLiD implementation instead samples from the VAE posterior and normalizes each channel with mean and standard deviation $0.5$.

\paragraph{Flickr30k shadow model.}
\label{sec:shadow-calibration}
After the same shuffle, zero-based rows $20{,}000$--$29{,}999$ provide $10{,}000$ shadow members and rows $30{,}000$--$31{,}782$ provide $1{,}783$ shadow non-members. Both groups are disjoint from the target-model training and evaluation images. The SD v1.4 shadow model uses the target's training recipe, effective batch size $16$, and $20{,}000$ updates. Its member and non-member features train the CLiD$_{\mathrm{vec}}$ classifier.

\subsection{Reference construction}
\label{sec:candidates}

Our reference families extend CLiD's caption reduction and embedding perturbations~\citep{zhai2024clid}. The language-model edits are inspired by SD-MIA's textual perturbations~\citep{qi2026sdmia} and use \texttt{gpt-4.1-mini}. Table~\ref{tab:prompt-families} summarizes their construction. The same procedure is applied to members and non-members. The expanded pools combine language-model edits, random captions, and the empty prompt. Case-only edits, word shuffling, and negation are excluded.

\begin{table}[htbp]
\centering
\small
\caption{Reference-caption families and their construction.}
\label{tab:prompt-families}
\begin{tabularx}{\linewidth}{@{}p{0.23\linewidth}X@{}}
\toprule
Prompt family & Construction \\
\midrule
\multicolumn{2}{@{}l}{\emph{Standard language-model rewrites}} \\
Token-level edits & Small wording changes, including one- or two-word synonym substitutions that preserve meaning. \\
Paraphrases & Rephrase the caption while preserving its meaning, including light and heavy rewrites with increasing changes in wording. \\
Style changes & Change how the caption is presented. \\
Semantic edits & Change depicted entities or actions, including variants that alter one, two, or three or more content words. \\
Combined edits & Change both wording and semantic content. \\
\midrule
\multicolumn{2}{@{}l}{\emph{Additional references}} \\
Random captions & Randomly sample $M$ captions of other images from the dataset's pooled caption corpus; exclude identical strings. \\
Empty prompt & Use the empty string $\varnothing$. \\
\bottomrule
\end{tabularx}
\end{table}

\FloatBarrier
\paragraph{Caption-rewriting prompts.}
The templates below share a system prompt. Here, \texttt{\{cap\}} is the original caption, \texttt{\{n\}} is the requested number of variants per family, and \texttt{H\_both} denotes combined edits.

\Needspace{6\baselineskip}
\noindent\textbf{System prompt}
\begin{Verbatim}[fontsize=\small,breaklines=true,breakautoindent=false,breaksymbolleft={},frame=single,framesep=3mm]
You rewrite image captions for a controlled experiment. Answer with STRICT JSON only, no prose. Keep every variant a natural image caption of similar length to the original.
\end{Verbatim}

\Needspace{12\baselineskip}
\noindent\textbf{User prompt: paraphrases, token-level, style, and semantic edits}
\begin{Verbatim}[fontsize=\small,breaklines=true,breakautoindent=false,breaksymbolleft={},frame=single,framesep=3mm]
Original caption: {cap}
Produce this JSON object exactly: {"para": [{n} paraphrases: SAME meaning, different wording, each clearly distinct from the others], "token": [{n} variants: change only 1-2 words, meaning essentially preserved], "style": [{n} variants: same content, add a style/medium/lighting modifier], "sem": [{n} variants: CHANGE one salient object or attribute so the described scene differs]}
\end{Verbatim}

\Needspace{12\baselineskip}
\noindent\textbf{User prompt: combined edits}
\begin{Verbatim}[fontsize=\small,breaklines=true,breakautoindent=false,breaksymbolleft={},frame=single,framesep=3mm]
Original caption: {cap}
Produce this JSON object exactly: {"H_both": [5 variants: HARD NEGATIVES that describe a clearly DIFFERENT scene. Each variant must (a) change THREE or more content words (objects, attributes, actions, places) so the scene is different, AND (b) lightly paraphrase the rest so that about half of the remaining words differ from the original wording. Keep the same sentence style and similar length; the five variants must be distinct from each other]}
\end{Verbatim}

\subsection{Reference pools and evaluation settings}
\label{sec:reference-set}

Table~\ref{tab:reference-counts} lists the number of references per family. Each reference slot contains one caption generated or sampled for each queried image using the same construction procedure. The main SD v1.4 evaluations use all eligible references; SD v2.1 and SD-XL use the reduced sets in Table~\ref{tab:sd21-configuration}, and LAION-MI uses random captions and the empty prompt. For each reference, residuals, prediction shifts, and loss gaps are computed from the same predictions and target noise. Separately evaluated caption families can use different noise draws. In that case, averaging across families also averages over more noise realizations.

\begin{table}[htbp]
\centering\small
\caption{References per image in the expanded pools, before zero-shift omissions.}
\label{tab:reference-counts}
\begin{tabular}{@{}lrrrr@{}}
\toprule
Reference family & Flickr30k & COCO & Pokemon & LAION-MI \\
\midrule
\multicolumn{5}{@{}l}{\emph{Standard language-model rewrites}} \\
Token-level edits & 6 & 8 & 14 & 6 \\
Paraphrases & 8 & 12 & 18 & 8 \\
Style changes & 2 & 4 & 4 & 2 \\
Semantic edits & 8 & 10 & 19 & 8 \\
Combined edits & 5 & 5 & 5 & 5 \\
\midrule
\multicolumn{5}{@{}l}{\emph{Additional references}} \\
Random captions & 12 & 6 & 6 & 6 \\
Empty prompt & 1 & 1 & 1 & 1 \\
\midrule
Total & 42 & 46 & 67 & 36 \\
\bottomrule
\end{tabular}
\end{table}

\paragraph{Zero-shift handling.}
\label{sec:zero-shifts}
Let $D$ be the number of elements in a noise prediction. We omit a state-reference pair when $\|\delta_{ij}\|_2^2/D\leq10^{-12}$ and average over all remaining pairs, as in Equation~\ref{eq:unified}. An image with no valid pairs receives score zero. The reference counts in Table~\ref{tab:reference-counts} are specified before this numerical check.

\paragraph{Configurations for additional analyses.}
\label{sec:diagnostic-references}
The scoring rules in Tables~\ref{tab:directional-readout} and~\ref{tab:signed-response} are compared using the default reference sets and diffusion states in Section~\ref{sec:default-configuration}. Each score averages its per-reference, per-state responses.

Table~\ref{tab:reference-state-readout} compares LoFIT and CLiD$_{\mathrm{th}}$ using either the complete pools in Table~\ref{tab:reference-counts} or the empty reference alone. Both scores use identical predictions over $48$ states on Flickr30k and LAION-MI and $16$ on COCO and Pokemon.

To track membership discrimination during fine-tuning, we evaluate successive model checkpoints using randomly sampled captions and the empty prompt as references for each dataset. Analyses of caption edits and Gaussian perturbations use the caption sets and query budgets specified for each experiment. All analyses use the same alignment score.

\Needspace{7\baselineskip}
\paragraph{SD v2.1 and SD-XL reference sets.}
\label{sec:sd21-configuration}
Table~\ref{tab:sd21-configuration} specifies the reduced reference sets used for the SD v2.1 and SD-XL results in Table~\ref{tab:main-summary}. Query counts use $TK(1+|\mathcal R|)$, reusing the original prediction across references at each state.

\begin{table}[htbp]
\centering\small
\caption{SD v2.1 and SD-XL reference counts and evaluation budgets.}
\label{tab:sd21-configuration}
\begin{tabular}{@{}lrrr@{}}
\toprule
Configuration & Flickr30k & COCO & Pokemon \\
\midrule
Combined edits & 5 & 5 & 5 \\
Semantic edits & 2 & 2 & 5 \\
Heavy paraphrases & 2 & 2 & 5 \\
Empty prompt & 1 & 1 & 1 \\
\midrule
Total references & 10 & 10 & 16 \\
Timesteps $T$ & 12 & 4 & 4 \\
Noise draws $K$ & 4 & 4 & 4 \\
Denoiser evaluations & 528 & 176 & 272 \\
\bottomrule
\end{tabular}
\end{table}

\subsection{Baseline implementations and calibration}
\label{sec:baseline-protocols}

This section specifies the baseline scoring procedures and the data used for calibration and classifier training.

\paragraph{Naive loss, PIA, and SecMI.}
Each attack uses its default fixed-timestep setting and one noise realization per image, with seed $0$. Predictions use the supplied caption without classifier-free guidance. Each method returns an image-level L2-distance score without classifier training or post-hoc calibration. Images and VAE latents follow the preprocessing above.

\paragraph{CLiD.}
We use the released character-clipping variant of CLiD~\citep{zhai2024clid}. Each image is evaluated under its original caption, three character-based caption thirds, and the empty prompt. One Gaussian noise draw per timestep is shared across conditions, giving $15$ denoiser queries per image. Evaluation uses sampled VAE latents and FP32 model weights. The features are the original-caption denoising MSE and four conditional loss gaps, each averaged over the three states.

For CLiD$_{\mathrm{th}}$, we apply \texttt{RobustScaler} to the original loss and negative mean gap before combining them with weights $1-\alpha$ and $\alpha$. Calibration follows the experiment-specific protocols below.

For the supplied-caption results on SD v1.4 and LAION-MI (Tables~\ref{tab:main-summary} and~\ref{tab:laion-summary}), we select oracle $\alpha\in\{0,0.1,\ldots,1\}$ to maximize AUROC using evaluation-pool membership labels. These entries give an oracle AUROC upper bound over the tested grid, rather than an independently calibrated attack. TPR is evaluated at the same selected $\alpha$. The selected values are $0.9$, $0.9$, $0.7$, and $0.6$ for Flickr30k, COCO, Pokemon, and LAION-MI, respectively. SD v2.1 uses fixed $\alpha=0.5$ with scaling fitted on the unlabeled evaluation pool, and SD-XL uses cross-fitting (Table~\ref{tab:clid-calibration}).

CLiD$_{\mathrm{vec}}$ uses the five features as inputs to the released XGBoost classifier. SD v1.4 Flickr30k uses shadow-model training; the remaining SD v1.4, SD v2.1, SD-XL, and LAION-MI settings use stratified two-fold cross-fitting on target-model features. We split member and non-member images separately, train on one fold, evaluate the other, and exchange the folds. We repeat this procedure with seeds $0$--$4$ and average the fold metrics across repetitions. For LAION-MI, we compute AUROC from pooled out-of-fold predictions. Cross-fitting uses labeled target examples in the training fold, with each scored image excluded from that fold. The classifier uses \texttt{n\_estimators=20}, \texttt{max\_depth=3}, \texttt{gamma=0.7}, \texttt{subsample=0.7}, \texttt{colsample\_bytree=0.7}, \texttt{reg\_alpha=1}, and \texttt{reg\_lambda=1}.

\begin{table}[htbp]
\centering\small
\caption{CLiD calibration and classifier configurations for Tables~\ref{tab:main-summary} and~\ref{tab:laion-summary}.}
\label{tab:clid-calibration}
\begin{tabularx}{\linewidth}{@{}p{.26\linewidth}XX@{}}
\toprule
Target setting & CLiD$_{\mathrm{th}}$ & CLiD$_{\mathrm{vec}}$ \\
\midrule
SD v1.4 Flickr30k & Oracle $\alpha=0.9$ & Shadow-trained classifier \\
SD v1.4 COCO, Pokemon & Oracle $\alpha=0.9$ and $0.7$ & Target two-fold cross-fitting \\
SD v2.1, all datasets & Fixed $\alpha=0.5$; unlabeled target-pool scaling & Target two-fold cross-fitting \\
SD-XL, all datasets & Target cross-fitting for scaler and $\alpha$ & Target two-fold cross-fitting \\
SD v1.5 LAION-MI & Oracle $\alpha=0.6$ & Target two-fold cross-fitting \\
\bottomrule
\end{tabularx}
\end{table}

\paragraph{CLiD$_{\mathrm{th}}$ on LoFIT's references and states.}
\label{sec:clid-matched-controls}
Table~\ref{tab:reference-state-readout} isolates the scoring rule by computing both methods from identical predictions. For each dataset and reference configuration, CLiD$_{\mathrm{th}}$ fits \texttt{RobustScaler} on the pooled evaluation examples and combines the scaled original loss and negative mean gap as above. We report fixed $\alpha=0.5$ and an oracle $\alpha$ selected to maximize evaluation AUROC. Each oracle row reports AUROC and TPR at that same $\alpha$. In Flickr30k/COCO/Pokemon/LAION-MI order, the oracle values are $1.0/0.9/0.8/0.7$ for all reference families and $0.9/0.9/0.7/0.7$ for the empty reference alone.

\paragraph{CLiD with BLIP-generated captions.}
\label{sec:clid-blip-calibration}
In Table~\ref{tab:generated-captions}, CLiD$_{\mathrm{th}}$ uses the released caption reductions and diffusion states with the BLIP-2 caption as the original condition. The same oracle protocol selects $\alpha=0.8$, $0.8$, $0.2$, and $0.4$ for Flickr30k, COCO, Pokemon, and LAION-MI, respectively. We report TPR at the same AUROC-maximizing $\alpha$.

\paragraph{CLiD calibration in the query-budget comparison.}
Figure~\ref{fig:budget} uses two-fold cross-fitting to calibrate CLiD$_{\mathrm{th}}$ on each dataset. Within each training fold, we fit the scaler and select $\alpha$ from $\{0,0.1,\ldots,1\}$ by AUROC, then score the held-out fold. We repeat the stratified two-fold split twice using a random generator initialized with seed $0$ and average the held-out metrics. This procedure is separate from the main-table calibration in Table~\ref{tab:clid-calibration}.

\paragraph{SD-MIA.}
We use the SD-MIA generation, caption-perturbation, and CLIP-scoring pipeline, with classifier-free guidance scale $6$ for generation.

\paragraph{PFAMI.}
We follow the released PFAMI statistical attack~\citep{fu2023pfami}. For text-to-image diffusion, we VAE-encode perturbed images at the target model's resolution and condition on the original caption with guidance scale $1$. Fine-tuned SD v1.4 and SD v2.1 targets use their corresponding pretrained backbones for calibration; LAION-MI uses the uncalibrated statistic. Evaluation costs $110$ denoiser queries per image, or $220$ with calibration.

\subsection{LAION-MI matching and population controls}
\label{sec:laion-distribution}
\label{sec:population-effects}

LAION-MI members originate from LAION-2B-en records included in Stable Diffusion training; non-members come from a later crawl~\citep{dubinski2024laionmi}. CLIP matching controls their image-text similarity discrepancy. We examine its effect on attack performance and the population differences that remain.

\paragraph{CLIP matching.}
We test whether CLIP matching removes class differences in image-text similarity. Before matching, mean similarities are $0.2698\pm0.0389$ for members and $0.2411\pm0.0526$ for non-members. Matching partitions their shared range into $20$ bins and subsamples each class to the smaller count per bin. Using CLIP similarity as the membership score, AUROC falls from $0.668$ to $0.503$ (Table~\ref{tab:laion-matching-audit}). This matching analysis retains $6{,}389$ examples per class; the final attack evaluation uses a separate matched population of $6{,}400$ examples per class.

\begin{table}[htbp]
\centering
\small
\caption{AUROC using CLIP image-text similarity as the membership score, before and after matching.}
\label{tab:laion-matching-audit}
\begin{tabular}{@{}lcc@{}}
\toprule
Membership score & Before matching & After matching \\
\midrule
CLIP image-text similarity & 0.668 & 0.503 \\
\bottomrule
\end{tabular}
\end{table}

\paragraph{Effect on SD-MIA.}
With the same $6{,}400$ members, matching reduces SD-MIA AUROC from $0.611$ against all $9{,}578$ non-members to $0.571$ against the $6{,}400$ retained non-members (Table~\ref{tab:laion-sdmia-populations}). The retained and excluded non-member groups are themselves distinguishable, showing that population differences can inflate apparent membership discrimination.

\begin{table}[htbp]
\centering
\small
\caption{SD-MIA AUROC before and after CLIP matching, with population controls on LAION-MI.}
\label{tab:laion-sdmia-populations}
\begin{tabular}{@{}lrrc@{}}
\toprule
Comparison & Group 1 size & Group 2 size & AUROC \\
\midrule
Members vs.\ all non-members (before) & 6,400 & 9,578 & 0.611 \\
Members vs.\ matched non-members (after) & 6,400 & 6,400 & 0.571 \\
Members vs.\ excluded non-members & 6,400 & 3,178 & 0.691 \\
Matched vs.\ excluded non-members & 6,400 & 3,178 & 0.635 \\
\bottomrule
\end{tabular}
\end{table}

\section{Evidence for the Directional Membership Signal}
\label{sec:alignment-analysis}

These analyses examine the position of the original prediction relative to denoising minima and identify which components of conditional predictions carry the membership signal. The loss-landscape and prediction-shift analyses use captions with varying degrees of wording and semantic modification, together with random captions and the empty prompt (Appendix~\ref{sec:diagnostic-references}).

\subsection{Loss minima along prediction shifts}
\label{sec:minimum-analysis}

\begin{figure}[htbp]
 \centering
 \includegraphics[width=\linewidth]{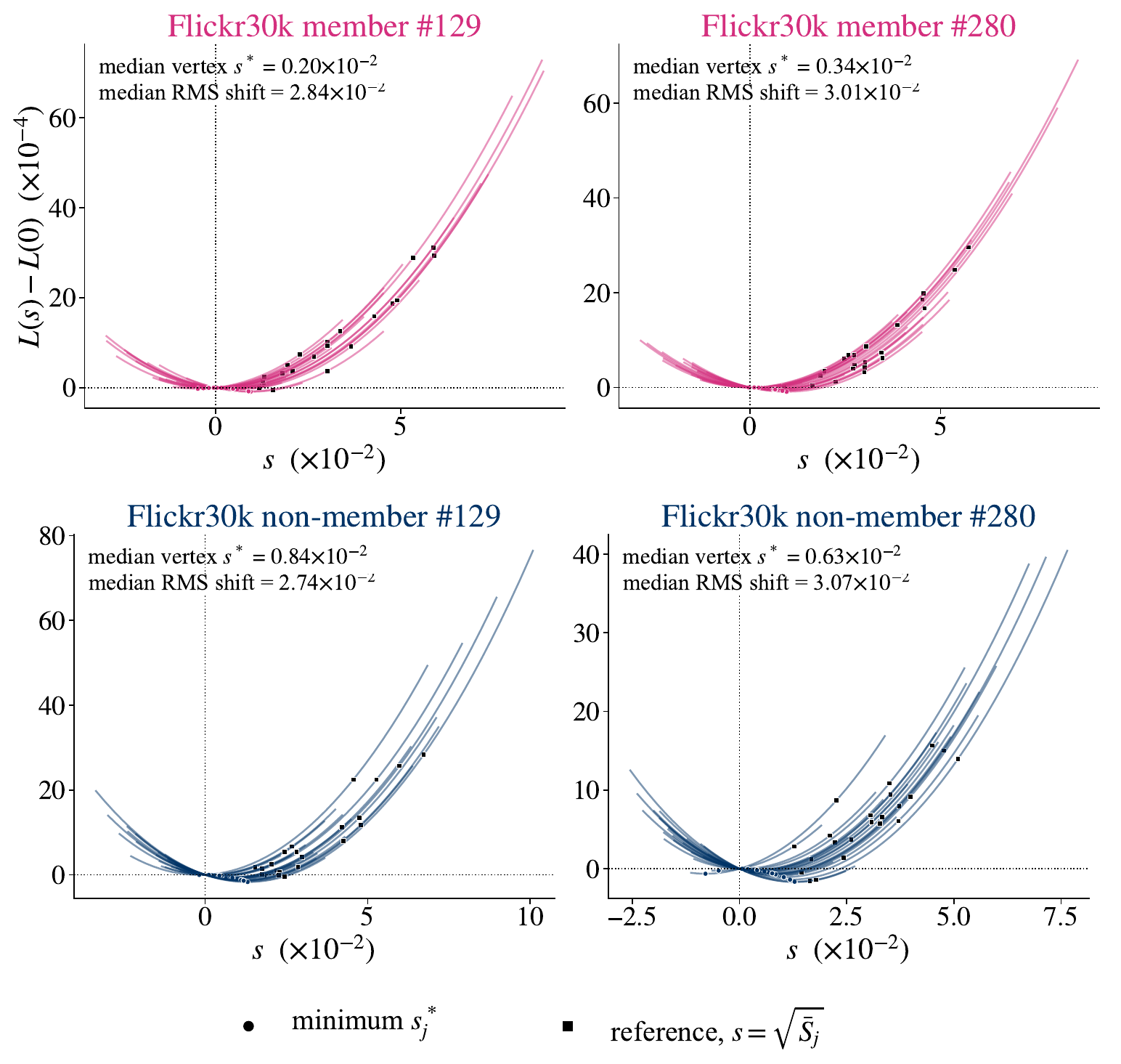}
 \caption{Loss landscapes for two Flickr30k members (top) and two non-members (bottom) under alternative reference captions. Each curve uses the state-averaged statistics for one reference. The original prediction is at $s=0$, and the vertical axis gives the mean loss change. Dots mark minima and squares reference predictions; curves include extrapolation. The RMS shift is $\sqrt{\bar S_j}$ (Equation~\ref{eq:directional-minimum}).}
 \label{fig:landscape-parabolas}
\end{figure}

\begin{figure}[htbp]
 \centering
 \includegraphics[width=\linewidth]{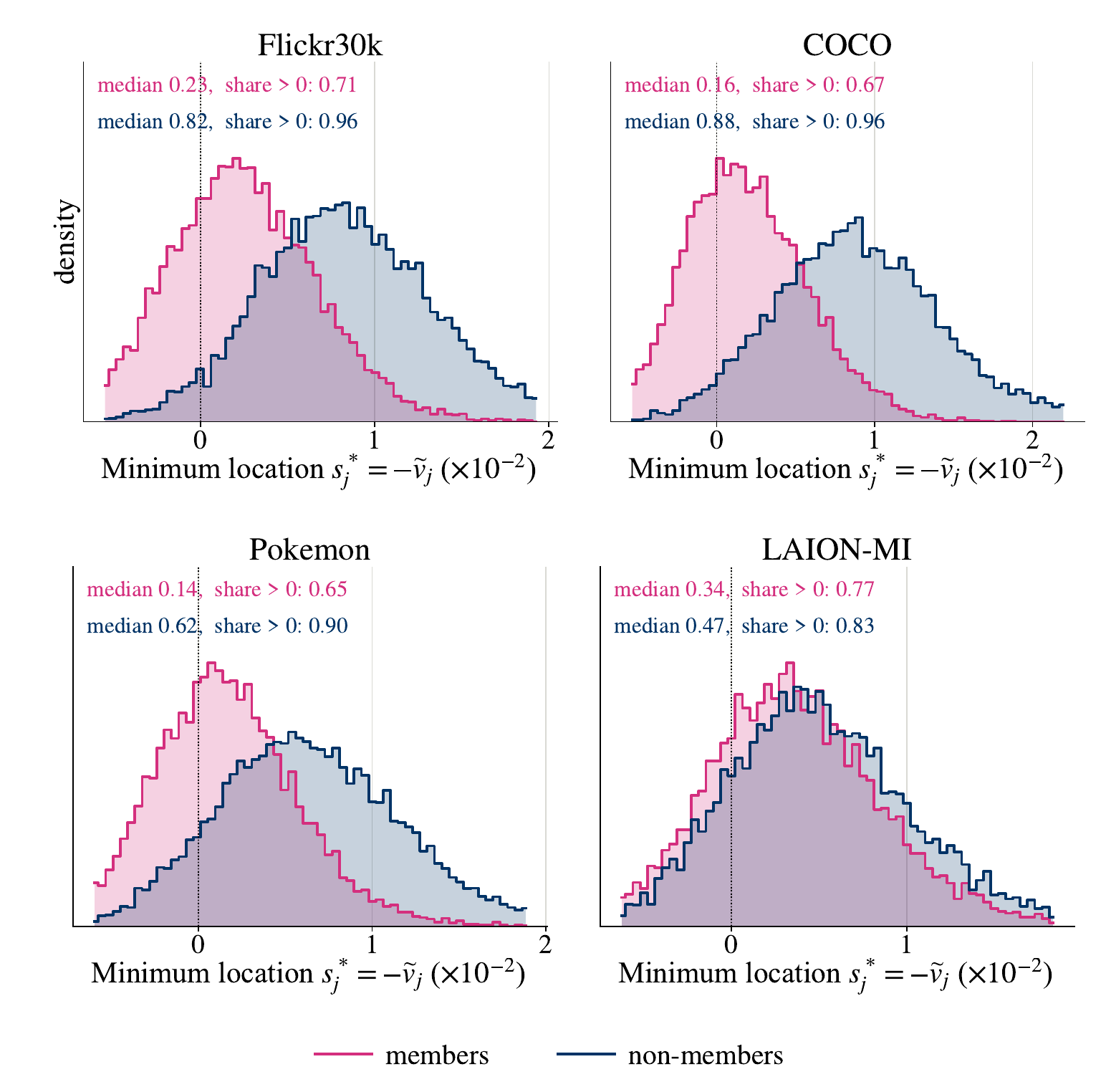}
 \caption{Distribution of minimum locations $s_j^*=-\widetilde v_j$ from the state-averaged loss curves. Each observation is an image-reference pair. Members' minima (magenta) concentrate closer to the original prediction at $s=0$ than non-members' minima (blue), with weaker separation on LAION-MI. Positive values indicate that the minimum lies toward the reference prediction.}
 \label{fig:landscape-parabola-minima}
\end{figure}

At fixed target checkpoints, we test whether members' original predictions lie closer to a denoising minimum along prediction shifts. Figures~\ref{fig:landscape-parabolas} and~\ref{fig:landscape-parabola-minima} show the loss curves and minimum locations.

This visualization uses a separate caption-edit experiment with $4$ timesteps and $4$ noise draws ($16$ states) per image, rather than the $48$-state Flickr30k configuration in the main evaluation. For each image-reference pair, $\bar X_j$ and $\bar S_j$ are the state averages of $\langle r_i,\delta_{ij}\rangle/D$ and $\|\delta_{ij}\|_2^2/D$, respectively, where $D$ is the prediction dimension. With the step expressed in root-mean-square prediction units, the mean loss change is
\begin{equation}
 \overline L_j(s)-\overline L_j(0)=2s\widetilde v_j+s^2,
 \qquad \widetilde v_j=\frac{\bar X_j}{\sqrt{\bar S_j}},
 \qquad s_j^*=-\widetilde v_j.
 \label{eq:directional-minimum}
\end{equation}
Here $s=0$ is the original prediction, the reference lies at $s=\sqrt{\bar S_j}$, and $s_j^*$ is the unconstrained minimum of the averaged curve. The curve coefficient $\widetilde v_j$ normalizes after state averaging; LoFIT instead normalizes each $v_{ij}$ before averaging.

Figure~\ref{fig:landscape-parabolas} illustrates the contrast on four Flickr30k examples: members' minima lie near the original prediction, while non-members' minima lie farther toward the references. Figure~\ref{fig:landscape-parabola-minima} shows this pattern across all four datasets. Median $s_j^*$ values for members/non-members are $0.23/0.82$ on Flickr30k, $0.16/0.88$ on COCO, and $0.14/0.62$ on Pokemon, all in units of $10^{-2}$. The difference is smaller on LAION-MI ($0.34/0.47$). Thus, the fine-tuned models show a consistent geometric signature: members' original predictions are closer to the minima along these reference-defined lines.

\subsection{Membership information in the loss-gap components}
\label{sec:residual-interaction}

\begin{figure}[htbp]
 \centering
 \includegraphics[width=.86\linewidth,trim=0 34 720.17 0,clip]{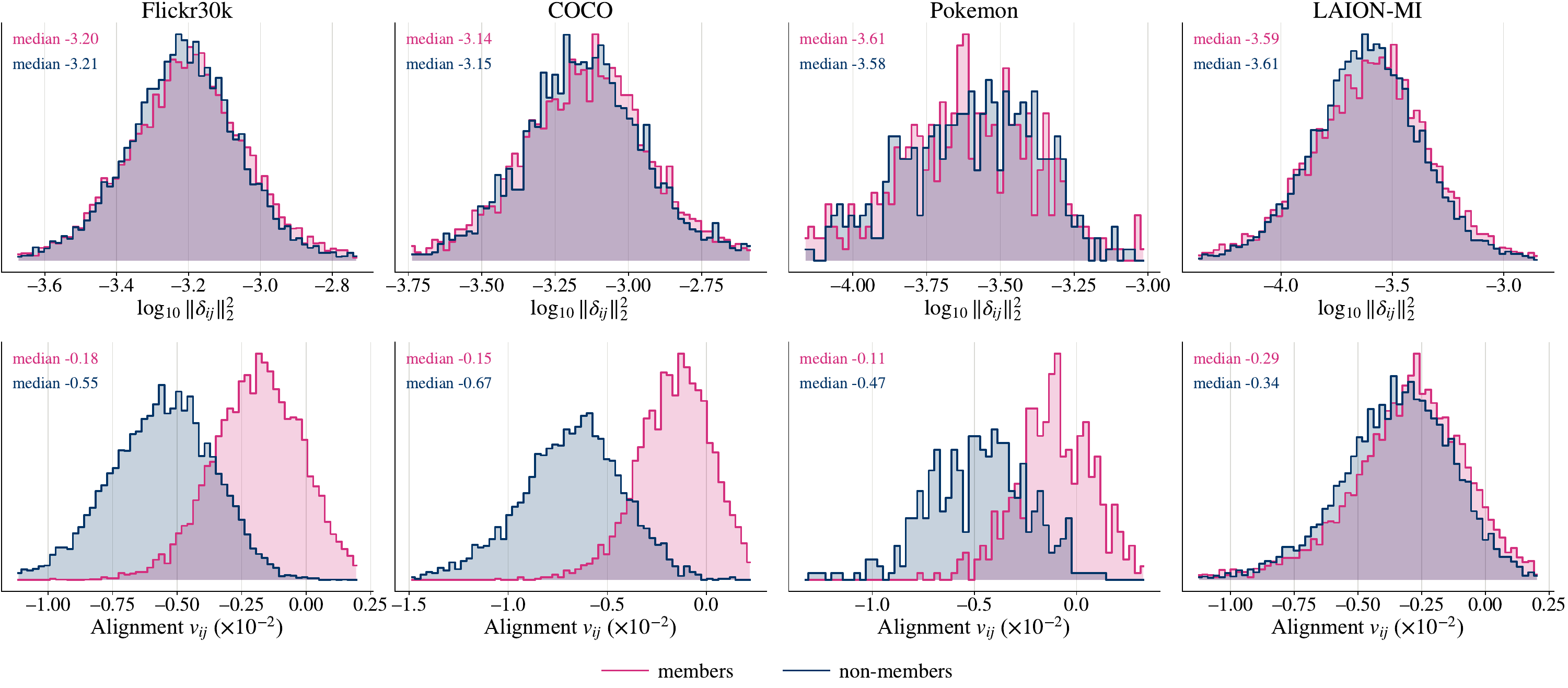}
 \par\vspace{2pt}
 \includegraphics[width=.86\linewidth,trim=720.17 34 0 0,clip]{figures/notation/shift_profile_types_hist.pdf}
 \par\vspace{2pt}
 \includegraphics[width=.42\linewidth,trim=550 0 550 600,clip]{figures/notation/shift_profile_types_hist.pdf}
 \caption{Conditional prediction statistics across four datasets. For each dataset, the upper panel shows $\log_{10}\|\delta_{ij}\|_2^2$ and the lower panel shows normalized alignment $\langle r_i,\delta_{ij}\rangle/\|\delta_{ij}\|_2$, the LoFIT response. Magenta and blue denote members and non-members, respectively. }
 \label{fig:shift-profile}
\end{figure}

We extend the loss-gap decomposition in Section~\ref{sec:signed-response} across all four datasets to examine which component carries membership information. For each dataset, Figure~\ref{fig:shift-profile} shows squared shift magnitudes in the upper panel and the LoFIT response $\langle r_i,\delta_{ij}\rangle/\|\delta_{ij}\|_2$ in the lower panel. Shift magnitudes largely overlap between the classes, while LoFIT responses are more negative for non-members. Separation is pronounced on the fine-tuned models and weaker on LAION-MI. Table~\ref{tab:directional-readout} tests these components on the full evaluation sets under the default reference configuration: mean shift magnitude remains near chance, while alignment provides stronger discrimination than the loss gap.

\subsection{Preserving the response sign improves discrimination}
\label{sec:signed-details}

We test whether retaining the sign of the alignment response improves membership discrimination. A negative $v_{ij}$ indicates that movement toward the reference prediction initially decreases denoising loss; a positive value indicates an increase. Taking the absolute value or square removes this distinction.

Table~\ref{tab:signed-response} compares these scoring rules on identical predictions using the settings in Appendix~\ref{sec:diagnostic-references}. Here $\mathbb E$ denotes averaging over references and diffusion states. The signed mean improves both AUROC and low-FPR TPR over the absolute and squared responses on all three fine-tuned models, and improves AUROC on LAION-MI. Preserving the direction of the initial loss change therefore strengthens the membership signal. We also test cosine normalization, which additionally divides each response by the residual norm. It lowers AUROC on the fine-tuned models and ties LoFIT on LAION-MI.

\begin{table}[htbp]
\centering
\small
\caption{Membership scores with and without the response sign, evaluated on identical predictions.}
\label{tab:signed-response}
\setlength{\tabcolsep}{2pt}
\begin{tabular*}{\linewidth}{@{\extracolsep{\fill}}lcccccccc@{}}
\toprule
& \multicolumn{2}{c}{Flickr30k} & \multicolumn{2}{c}{COCO} & \multicolumn{2}{c}{Pokemon} & \multicolumn{2}{c}{LAION-MI} \\
\cmidrule(lr){2-3}\cmidrule(lr){4-5}\cmidrule(lr){6-7}\cmidrule(lr){8-9}
Score & AUROC & \shortstack{TPR@1\%\\FPR} & AUROC & \shortstack{TPR@1\%\\FPR} & AUROC & \shortstack{TPR@1\%\\FPR} & AUROC & \shortstack{TPR@5\%\\FPR} \\
\midrule
$\mathbb E[v]$ (ours) & \textbf{0.935} & \textbf{0.454} & \textbf{0.954} & \textbf{0.478} & \textbf{0.896} & \textbf{0.272} & \textbf{0.602} & \textbf{0.101} \\
$-\mathbb E[|v|]$ & 0.893 & 0.183 & 0.915 & 0.246 & 0.837 & 0.062 & 0.595 & 0.086 \\
$-\mathbb E[v^2]$ & 0.886 & 0.184 & 0.910 & 0.216 & 0.835 & 0.087 & 0.595 & 0.088 \\
$\mathbb E[v/\|r\|_2]$ & 0.917 & 0.377 & 0.944 & 0.410 & 0.891 & 0.255 & 0.602 & 0.101 \\
\bottomrule
\end{tabular*}
\end{table}

\section{Choosing Informative Conditioning Directions}
\label{sec:direction-design}

This section examines how reference selection and perturbation design affect LoFIT's membership discrimination. We study which references provide strong and complementary signals, compare conditioning changes with alternative perturbations, and assess how reference selection changes when the original caption is unavailable.

\subsection{Reference quality and complementarity}
\label{sec:alignment-aggregation}

To guide reference selection, we examine how well individual reference captions distinguish members from non-members and whether combining caption families improves discrimination.

\paragraph{Reference quality at equal cost.}
We evaluate each reference independently at equal query cost to separate reference quality from the benefit of averaging. Table~\ref{tab:full-family-ablation} reports the mean and standard deviation of AUROC across reference slots within each family. We use the expanded reference pools in Appendix~\ref{sec:reference-set}, the full evaluation sets, and the diffusion-state configurations in Section~\ref{sec:default-configuration}, requiring $2TK$ denoiser evaluations per image. The empty prompt contributes a single slot.

\begin{table}[htbp]
\centering\small
\caption{Single-reference LoFIT AUROC (mean $\pm$ standard deviation across slots) at equal query cost. Bold marks the highest mean per dataset.}
\label{tab:full-family-ablation}
\setlength{\tabcolsep}{4pt}
\begin{tabular}{@{}lcccc@{}}
\toprule
Reference family & Flickr30k & COCO & Pokemon & LAION-MI \\
\midrule
Token-level edits & $0.642\pm0.022$ & $0.679\pm0.026$ & $0.655\pm0.030$ & $0.515\pm0.008$ \\
Paraphrases & $0.737\pm0.027$ & $0.756\pm0.029$ & $0.701\pm0.041$ & $0.517\pm0.009$ \\
Style changes & $0.650\pm0.033$ & $0.700\pm0.024$ & $0.638\pm0.015$ & $0.504\pm0.010$ \\
Semantic edits & $0.743\pm0.064$ & $0.793\pm0.058$ & $0.738\pm0.047$ & $0.529\pm0.017$ \\
Combined edits & $0.880\pm0.002$ & $0.909\pm0.007$ & $\mathbf{0.810}\pm0.014$ & $0.569\pm0.006$ \\
\midrule
Random captions & $0.836\pm0.009$ & $0.819\pm0.011$ & $0.740\pm0.017$ & $0.574\pm0.007$ \\
Empty prompt & $\mathbf{0.900}$ & $\mathbf{0.914}$ & $0.781$ & $\mathbf{0.606}$ \\
\bottomrule
\end{tabular}
\end{table}

The empty prompt is strongest on Flickr30k, COCO, and LAION-MI; combined edits lead on Pokemon and among language-model rewrites on every dataset. Semantic edits vary more across references; on COCO, their mean single-reference AUROC is $0.793$, compared with $0.819$ for random captions.

Combined edits also outperform random captions when both use four references. For each family, we randomly sample four references and evaluate AUROC, repeating this procedure $25$ times. The mean AUROC is $0.919$ versus $0.894$ on Flickr30k and $0.940$ versus $0.872$ on COCO. Averaging also benefits semantic edits: on COCO, their mean AUROC increases from $0.793$ with one reference to $0.897$ with four.

\subsection{Caption-based and Gaussian conditioning probes}
\label{sec:random-directions}

We compare the membership information supplied by individual caption and Gaussian references, then examine their combination. We use the full evaluation sets and diffusion-state grids in Table~\ref{tab:full-family-ablation}. Each reference is scored independently by averaging alignment responses over states. We report the mean and standard deviation of AUROC across references, at a cost of $2TK$ queries per image per reference.

Gaussian probes query $e_0+hU_j$, where $e_0\in\mathbb R^{77\times768}$ is the original caption's conditioning tensor and $U_j$ is Gaussian noise normalized to unit Frobenius norm. We sample four Gaussian directions per image at each $h\in\{40,80,160\}$ and reuse each direction across that image's diffusion states. Table~\ref{tab:reference-directions-full} averages the four single-direction AUROCs at each scale and reports their standard deviation where available; caption-family counts appear in Table~\ref{tab:reference-counts}. The denoiser converts each conditioning perturbation into a prediction shift, which is scored using the same alignment formula as a caption reference. These probes require direct access to the conditioning embedding interface, beyond the text-condition queries assumed by LoFIT.

We match the caption and Gaussian collections by image and timestep/noise-draw index. Their original-caption denoising losses agree exactly on all four datasets.

\begin{table}[htbp]
\centering
\small
\caption{Single-reference AUROC for caption and Gaussian probes. Entries show means and standard deviations where available.}
\label{tab:reference-directions-full}
\setlength{\tabcolsep}{4pt}
\begin{tabular}{@{}lcccc@{}}
\toprule
Reference & Flickr30k & COCO & Pokemon & LAION-MI \\
\midrule
\multicolumn{5}{@{}l}{\emph{Caption references}} \\
Empty prompt & $0.900$ & $0.914$ & $0.781$ & $0.606$ \\
Combined edits & $0.880\pm0.002$ & $0.909\pm0.007$ & $0.810\pm0.014$ & $0.569$ \\
Random captions & $0.836\pm0.009$ & $0.819\pm0.011$ & $0.740\pm0.017$ & $0.574$ \\
Semantic edits & $0.743\pm0.064$ & $0.793\pm0.058$ & $0.738\pm0.047$ & $0.529$ \\
Paraphrases & $0.737\pm0.027$ & $0.756\pm0.029$ & $0.701\pm0.041$ & $0.517$ \\
Style changes & $0.650\pm0.033$ & $0.700\pm0.024$ & $0.638\pm0.015$ & $0.504$ \\
Token-level edits & $0.642\pm0.022$ & $0.679\pm0.026$ & $0.655\pm0.030$ & $0.515$ \\
\midrule
\multicolumn{5}{@{}l}{\emph{Gaussian conditioning probes}} \\
$h=40$ & $0.688\pm0.002$ & $0.731\pm0.004$ & $0.646\pm0.025$ & $0.520$ \\
$h=80$ & $0.806\pm0.001$ & $0.863\pm0.004$ & $0.761\pm0.014$ & $0.534$ \\
$h=160$ & $0.854\pm0.001$ & $0.903\pm0.003$ & $0.805\pm0.009$ & $0.534$ \\
\bottomrule
\end{tabular}
\end{table}

On the fine-tuned models, a single Gaussian direction at $h=160$ outperforms a random caption on average and approaches combined edits, particularly on COCO and Pokemon. This ranking differs on pretrained LAION-MI: Gaussian probes yield AUROC $0.520$--$0.534$, compared with $0.574$ for random captions and $0.606$ for the empty prompt. These single-reference results favor Gaussian probes on the fine-tuned models and caption references on LAION-MI at the tested perturbation scales.

\paragraph{Combining captions and Gaussian probes.}
Adding Gaussian probes to the caption pool improves membership discrimination on all three fine-tuned models (Table~\ref{tab:caption-gaussian-combination}). We pool alignment responses from the complete caption reference sets in Table~\ref{tab:reference-counts} and eight additional Gaussian probes, four at $h=80$ and four at $h=160$. These combined configurations use more queries than the caption-only configurations. AUROC increases from $0.935$ to $0.946$ on Flickr30k, $0.954$ to $0.966$ on COCO, and $0.896$ to $0.910$ on Pokemon. TPR at $1\%$ FPR rises from $0.454$ to $0.501$ on Flickr30k and from $0.478$ to $0.565$ on COCO. On LAION-MI, combining the full caption pool with Gaussian probes changes AUROC from $0.573$ to $0.575$.

\begin{table}[htbp]
\centering
\small
\caption{AUROC after adding Gaussian probes to the full caption pool.}
\label{tab:caption-gaussian-combination}
\setlength{\tabcolsep}{6pt}
\begin{tabular}{@{}lcccc@{}}
\toprule
References & Flickr30k & COCO & Pokemon & LAION-MI \\
\midrule
Caption pool & 0.935 & 0.954 & 0.896 & 0.573 \\
Caption + Gaussian & 0.946 & 0.966 & 0.910 & 0.575 \\
\bottomrule
\end{tabular}
\end{table}

\raggedbottom
\subsection{Membership inference with generated captions}
\label{sec:generated-captions}

We evaluate image-membership inference when the original caption is unavailable. A BLIP-2 OPT-6.7b~\citep{li2023blip2} caption, generated by greedy decoding, replaces $c_0$ throughout scoring and reference construction. Membership labels indicate whether the image was used for training. All results use the full evaluation populations. LoFIT and the loss-gap controls use the timestep grids in Section~\ref{sec:settings} and four noise draws.

\paragraph{Reference construction.}
Caption references comprise randomly sampled BLIP-2 captions of other images in the pool, the empty prompt, and language-model rewrites of the generated caption. Gaussian references perturb its conditioning tensor as $e_{\mathrm B}+hU_j$, where $e_{\mathrm B}$ encodes the BLIP-2 caption and $U_j$ is normalized Gaussian noise, as defined in Appendix~\ref{sec:random-directions}. We test three directions at each $h\in\{40,80,160\}$ and average alignment responses over directions and states; the nine-direction configuration pools all three scales. The loss-gap baseline averages conditional loss differences on the same predictions used by LoFIT. A single-scale Gaussian configuration requires $192$ denoiser evaluations per image on Flickr30k and LAION-MI, and $64$ on COCO and Pokemon, excluding caption generation.

Official CLiD uses the released evaluation configuration with oracle $\alpha$, as specified in Appendix~\ref{sec:clid-blip-calibration}.

\begin{table}[H]
\centering
\caption{Image-membership inference with BLIP-2 captions. Bold marks the best value per metric.}
\label{tab:generated-captions}
\small
\setlength{\tabcolsep}{1.5pt}
\begin{tabular*}{\linewidth}{@{\extracolsep{\fill}}p{0.26\linewidth}cccccccc@{}}
\toprule
& \multicolumn{2}{c}{Flickr30k} & \multicolumn{2}{c}{COCO} & \multicolumn{2}{c}{Pokemon} & \multicolumn{2}{c}{LAION-MI} \\
\cmidrule(lr){2-3}\cmidrule(lr){4-5}\cmidrule(lr){6-7}\cmidrule(lr){8-9}
Score and references & AUROC & \shortstack{TPR@1\%\\FPR} & AUROC & \shortstack{TPR@1\%\\FPR} & AUROC & \shortstack{TPR@1\%\\FPR} & AUROC & \shortstack{TPR@5\%\\FPR} \\
\midrule
\multicolumn{9}{@{}l}{\emph{Official CLiD}} \\
CLiD$_{\mathrm{th}}$, oracle $\alpha$ & 0.603 & 0.031 & 0.650 & 0.044 & 0.576 & 0.041 & 0.535 & 0.084 \\
\midrule
\multicolumn{9}{@{}l}{\emph{Loss-based controls}} \\
Naive denoising loss & 0.522 & 0.014 & 0.525 & 0.018 & 0.554 & 0.026 & 0.521 & 0.071 \\
Mean loss gap, empty & 0.627 & 0.063 & 0.673 & 0.059 & 0.523 & 0.041 & 0.536 & 0.081 \\
Mean loss gap, random + empty & 0.587 & 0.033 & 0.619 & 0.039 & 0.567 & 0.046 & 0.527 & 0.087 \\
\midrule
\multicolumn{9}{@{}l}{\emph{LoFIT with caption references}} \\
Empty only & 0.680 & 0.073 & 0.763 & 0.103 & 0.530 & 0.007 & \textbf{0.564} & \textbf{0.092} \\
Random + empty & 0.641 & 0.029 & 0.669 & 0.027 & 0.621 & 0.034 & 0.546 & 0.076 \\
Language-model rewrites & 0.684 & 0.070 & 0.752 & 0.091 & 0.605 & 0.062 & 0.545 & 0.077 \\
\midrule
\multicolumn{9}{@{}l}{\emph{LoFIT with Gaussian references}} \\
$h=40$ & 0.749 & 0.081 & 0.800 & 0.106 & 0.669 & 0.046 & 0.535 & 0.063 \\
$h=80$ & 0.838 & 0.190 & 0.894 & 0.236 & 0.782 & 0.130 & 0.544 & 0.066 \\
$h=160$ & \textbf{0.842} & \textbf{0.214} & \textbf{0.904} & \textbf{0.264} & \textbf{0.802} & \textbf{0.139} & 0.532 & 0.068 \\
All nine directions & 0.837 & 0.192 & 0.896 & 0.252 & 0.777 & 0.111 & 0.540 & 0.066 \\
Mean loss gap, same nine & 0.618 & 0.038 & 0.648 & 0.057 & 0.600 & 0.036 & 0.494 & 0.054 \\
\midrule
\multicolumn{9}{@{}l}{\emph{LoFIT with mixed references}} \\
Random + empty + Gaussian & 0.755 & 0.096 & 0.849 & 0.193 & 0.761 & 0.072 & 0.552 & 0.081 \\
All captions + Gaussian & 0.749 & 0.117 & 0.830 & 0.186 & 0.717 & 0.087 & 0.554 & 0.078 \\
\bottomrule
\end{tabular*}
\end{table}

\begin{samepage}
\paragraph{Alignment remains informative with generated captions.}
With random captions and the empty reference, LoFIT exceeds official CLiD$_{\mathrm{th}}$ with oracle $\alpha$ in AUROC across all four datasets (Table~\ref{tab:generated-captions}). On identical predictions, it also outperforms the loss-gap baseline in AUROC with the empty reference alone, random captions plus empty, and the nine Gaussian directions.

\end{samepage}

With BLIP-generated captions, Gaussian probes outperform caption references on the fine-tuned models. At $h=160$, AUROC reaches $0.842$, $0.904$, and $0.802$ on Flickr30k, COCO, and Pokemon, respectively. Using the supplied captions yields higher AUROC (Table~\ref{tab:main-summary}). Within the BLIP setting, combining caption references with Gaussian probes lowers AUROC relative to using the nine Gaussian directions alone on these datasets. On LAION-MI, the empty reference yields AUROC $0.564$, above the caption mixtures and Gaussian probes in Table~\ref{tab:generated-captions}. These results show that the most informative references depend on the available query caption.

\clearpage

\section{Membership Inference on Video Diffusion Models}
\label{sec:video-results}

We evaluate whether LoFIT's conditional membership signal extends to video generation using CogVideoX-2b~\citep{yang2025cogvideox} and Wan2.1~\citep{wan2025video} on Anime-Shooter~\citep{qiu2025animeshooter} and OpenVid~\citep{nan2025openvid}.

\paragraph{Training.}
We fully fine-tune both models for $200$ iterations on Anime-Shooter. On OpenVid, we fully fine-tune CogVideoX-2b for $4{,}000$ iterations and Wan2.1 for $7{,}000$ iterations. All runs use batch size $64$. Membership denotes inclusion in the fine-tuning data.

\paragraph{Evaluation.}
On Anime-Shooter, we evaluate $500$ members for each backbone, with $4{,}000$ non-members for CogVideoX-2b and $3{,}998$ for Wan2.1. On OpenVid, we evaluate $1{,}000$ members and $4{,}000$ non-members for both backbones. We report AUROC and TPR at $5\%$ FPR (Table~\ref{tab:video-results}).

For CogVideoX, we use $12$ timesteps $t\in\{80,160,\ldots,960\}$ and construct noisy video latents. For Wan2.1, the same nominal timesteps map to noise levels $\sigma=t/1000\in\{0.08,0.16,\ldots,0.96\}$, with
\begin{equation}
 x_\sigma=(1-\sigma)x_0+\sigma\epsilon,
 \label{eq:video-wan-noising}
\end{equation}
where $x_0$ is the clean video latent and $\epsilon\sim\mathcal N(0,I)$. Wan2.1 losses and alignment responses are evaluated in velocity-prediction space. For each video and state, we draw one noise tensor using a fixed random seed and share it across conditions, so the original and reference predictions use the same noisy input.

The original condition is the clip's training caption, and the sole reference is the empty prompt. LoFIT uniformly averages the responses over all $12$ predefined states, requiring $24$ denoiser forward evaluations per video ($12$ states $\times$ $2$ conditions). PIA and SecMI use $13$ and $6$ forward evaluations per video, respectively.

\begin{table}[!htb]
\centering\small
\caption{Video membership inference on fine-tuned CogVideoX-2b and Wan2.1. Bold marks the best value per model, dataset, and metric.}
\label{tab:video-results}
\setlength{\tabcolsep}{3pt}
\begin{tabular}{@{}lcccccccc@{}}
\toprule
& \multicolumn{4}{c}{CogVideoX-2b} & \multicolumn{4}{c}{Wan2.1} \\
\cmidrule(lr){2-5}\cmidrule(lr){6-9}
& \multicolumn{2}{c}{Anime-Shooter} & \multicolumn{2}{c}{OpenVid} & \multicolumn{2}{c}{Anime-Shooter} & \multicolumn{2}{c}{OpenVid} \\
\cmidrule(lr){2-3}\cmidrule(lr){4-5}\cmidrule(lr){6-7}\cmidrule(lr){8-9}
Method & AUROC & \shortstack{TPR@5\%\\FPR} & AUROC & \shortstack{TPR@5\%\\FPR} & AUROC & \shortstack{TPR@5\%\\FPR} & AUROC & \shortstack{TPR@5\%\\FPR} \\
\midrule
Naive loss & 0.580 & 0.122 & 0.504 & 0.044 & 0.515 & 0.072 & 0.502 & 0.056 \\
PIA & 0.547 & 0.090 & 0.502 & 0.052 & 0.504 & 0.052 & 0.490 & 0.050 \\
SecMI & 0.508 & 0.074 & 0.591 & 0.108 & 0.488 & 0.068 & 0.493 & 0.046 \\
CLiD$_{\mathrm{th}}$ ($\alpha=0.5$) & 0.870 & 0.530 & 0.594 & 0.114 & 0.564 & 0.074 & 0.547 & 0.072 \\
\textbf{LoFIT (ours)} & \textbf{0.948} & \textbf{0.692} & \textbf{0.656} & \textbf{0.116} & \textbf{0.610} & \textbf{0.112} & \textbf{0.604} & \textbf{0.105} \\
\bottomrule
\end{tabular}
\end{table}

\paragraph{Results.}
LoFIT achieves the highest AUROC and TPR across all four fine-tuning settings. On CogVideoX-2b, it reaches AUROC $0.948$ on Anime-Shooter and $0.656$ on OpenVid, compared with $0.870$ and $0.594$ for CLiD$_{\mathrm{th}}$. On Wan2.1, LoFIT achieves AUROC $0.610$ and TPR $0.112$ on Anime-Shooter, exceeding CLiD$_{\mathrm{th}}$ at $0.564$ and $0.074$. On OpenVid, it achieves AUROC $0.604$ and TPR $0.105$, compared with $0.547$ and $0.072$ for CLiD$_{\mathrm{th}}$. These results extend the conditional membership signal across two video-generation backbones.

\FloatBarrier
\end{document}